\documentclass{article}
\usepackage{adjustbox}
\usepackage{diagbox} 
\usepackage[table]{xcolor}
\usepackage{algorithm}
\newtheorem{theorem}{Theorem}
\newtheorem{lemma}{Lemma}
    \PassOptionsToPackage{numbers, compress}{natbib}

\usepackage[final]{neurips_2022}
\usepackage{algorithmic}
\usepackage[utf8]{inputenc} 
\usepackage[T1]{fontenc}    
\usepackage[colorlinks = true,
            linkcolor = blue,
            urlcolor  = blue,
            citecolor = blue,
            anchorcolor = blue]{hyperref}

\usepackage{url}            
\usepackage{booktabs}       
\usepackage{amsfonts}       
\usepackage{nicefrac}       
\usepackage{microtype}      
\usepackage{xcolor}         
\usepackage{soul}
\usepackage{url}
\usepackage{amsmath}
\usepackage{booktabs}
\usepackage{listings}
\usepackage{url}
\usepackage{subcaption}

\usepackage{color}
\definecolor{codegreen}{rgb}{0,0.5,0}
\definecolor{codeblue}{rgb}{0,0,0.9}
\definecolor{codeblues}{rgb}{0,0,0.4}
\definecolor{codegray2}{rgb}{0.4,0.4,0.4}
\definecolor{codegray}{rgb}{0.9,0.9,0.9}
\definecolor{codepurple}{rgb}{0.58,0,0.82}
\definecolor{backcolour}{rgb}{0.95,0.95,0.92}
\definecolor{backcolour2}{rgb}{0.9,0.9,0.9}
\definecolor{codered}{rgb}{0.5,0,0}
\definecolor{textcodered}{rgb}{0.05,0.05,0.05}
\definecolor{palegray}{rgb}{0.98,0.98,0.99}
\usepackage{textcomp}

\usepackage[T1]{fontenc}    
\usepackage{url}            
\usepackage{booktabs}       
\usepackage{amsfonts}       
\usepackage{nicefrac}       
\usepackage{microtype}      
\usepackage{graphicx}
\usepackage{epstopdf}
\usepackage{epsfig}
\usepackage{xspace}
\usepackage{natbib}
\usepackage{setspace}

\makeatletter  
\newif\if@restonecol  
\makeatother  
\usepackage{multirow}
\usepackage{hyperref}

\newcommand{\modelname}{DOMINO\xspace}

\newcommand*{\mi}{\ensuremath{I}}
\newcommand*{\imgintext}[1]{%
  \raisebox{-.3\baselineskip}{%
    \centering\includegraphics[
      height=\baselineskip,
      width=\baselineskip,
      keepaspectratio,
    ]{#1}%
  }%
}
\newcommand{\expect}{\mathbb{E}}
\newcommand{\x}{\ensuremath{x}}
\newcommand{\xt}{y}

\usepackage[utf8]{inputenc} 
\usepackage[T1]{fontenc}    
\usepackage{hyperref}       
\usepackage{url}            
\usepackage{booktabs}       
\usepackage{amsfonts}       
\usepackage{nicefrac}       
\usepackage{microtype}      
\usepackage{xcolor}         

\title{Decomposed Mutual Information Optimization for Generalized Context in Meta-Reinforcement Learning }

\author{%
  Yao Mu\\
  The University of Hong Kong\\
  \texttt{\small{muyao@connect.hku.hk}} \\
  \And
  Yuzheng Zhuang   \\
  Huawei Noah's Ark Lab\\
  \texttt{\small{zhuangyuzheng@huawei.com}} \\
  \And
  Fei Ni   \\
  Tianjin University\\
  \texttt{\small{fei\_ni@tju.edu.cn}} \\
  \And
  Bin Wang \\
  Huawei Noah's Ark Lab\\
  \texttt{\small{wangbin158@huawei.com}} \\
  \And
  Jianyu Chen \\
  Tsinghua University\\
  \texttt{{jianyuchen@tsinghua.edu.cn}} \\
   \AND
  Jianye Hao \\
  Huawei Noah's Ark Lab\\
  \texttt{\small{haojianye@huawei.com}} \\
    \And
  Ping Luo \thanks{Ping Luo is the corresponding author. Yao Mu and Fei Ni conducted this work
during the internship in Huawei Noah’s Ark Lab. }\\
  The University of Hong Kong\\
  \texttt{\small{pluo@cs.hku.hk}} \\
}

\begin{document}

\maketitle

\begin{abstract}
Adapting to the changes in transition dynamics is essential in robotic applications. By learning a conditional policy with a compact context, context-aware meta-reinforcement learning provides a flexible way to adjust behavior according to dynamics changes. However, in real-world applications, the agent may encounter complex dynamics changes. Multiple confounders can   influence the transition dynamics, making it challenging to infer accurate context for decision-making.
This paper addresses such a challenge by \textbf{D}ec\textbf{O}mposed \textbf{M}utual \textbf{IN}formation \textbf{O}ptimization (\modelname) for context learning, which explicitly learns a disentangled context to maximize the mutual information between the context and historical trajectories, while minimizing the state transition prediction error.
Our theoretical analysis shows that \modelname can overcome the underestimation of the mutual information caused by multi-confounded challenges via learning disentangled context and reduce the demand for the number of samples collected in various environments.
Extensive experiments show that the context learned by \modelname benefits both model-based and model-free reinforcement learning algorithms for dynamics generalization in terms of sample efficiency and performance in unseen environments. Open-sourced code is released on our \href{https://sites.google.com/view/dominorl/}{homepage}.
\end{abstract}

\section{Introduction}
\label{submission}
 {Dynamics generalization in deep reinforcement learning (RL) investigates the problem of training a RL agent in a few kinds of environments and adapting across unseen system dynamics or structures, such as different physical parameters or robot mythologies. }
Meta-Reinforcement Learning (Meta-RL) has been proposed to tackle the problem by training on a range of tasks, and fast adapting to a new task with the learned prior knowledge. 
However, training in meta-RL requires orders of magnitudes more samples than single-task RL since the agent not only has to learn to infer the change of environment but also has to learn the corresponding policies. 
Context-aware meta-RL methods take a step further and show promising potential to capture local dynamics explicitly by learning an
additional context vector from historical trajectories  ~\cite{DBLP:conf/icml/RakellyZFLQ19, DBLP:conf/iclr/FakoorCSS20,lee2020context,seo2020trajectory}. The historical trajectories are sampled from the joint distribution of multiple confounders, which are the key factors that cause the dynamics changes.  Accordingly, if multiple confounders affect the dynamics simultaneously, the state transition distribution will become highly multi-modal, leading to challenges in extracting accurate context.

\begin{figure}[t]
\vspace{-15pt}
        \label{fig:INTRO}
     \centering
     \begin{subfigure}[b]{0.63\textwidth}
         \centering
         \includegraphics[width=\textwidth]{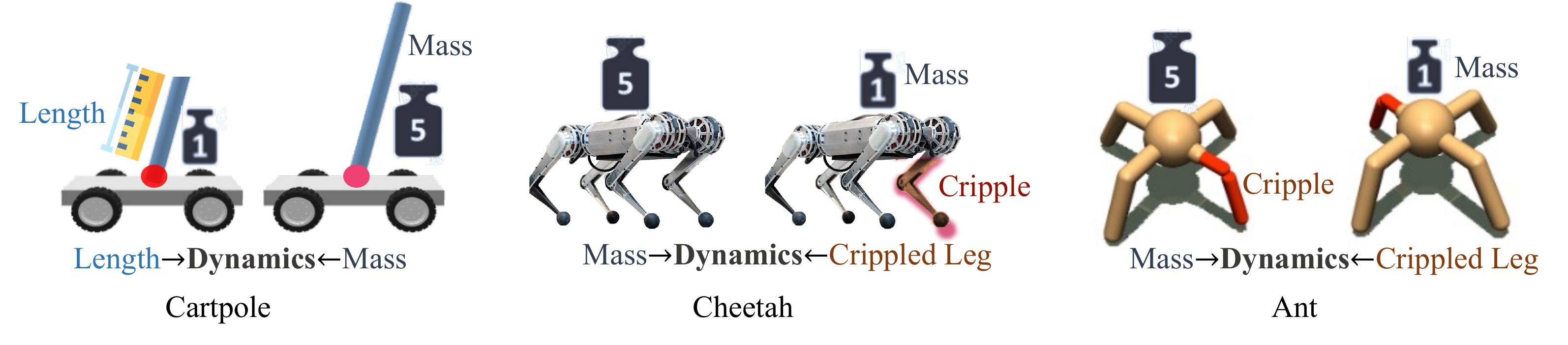}
         \caption{Example of the   multi-confounded environments}
         \label{fig:Dynamic Generalization}
     \end{subfigure}
     \begin{subfigure}[b]{0.36\textwidth}
         \centering
         \includegraphics[width=\textwidth]{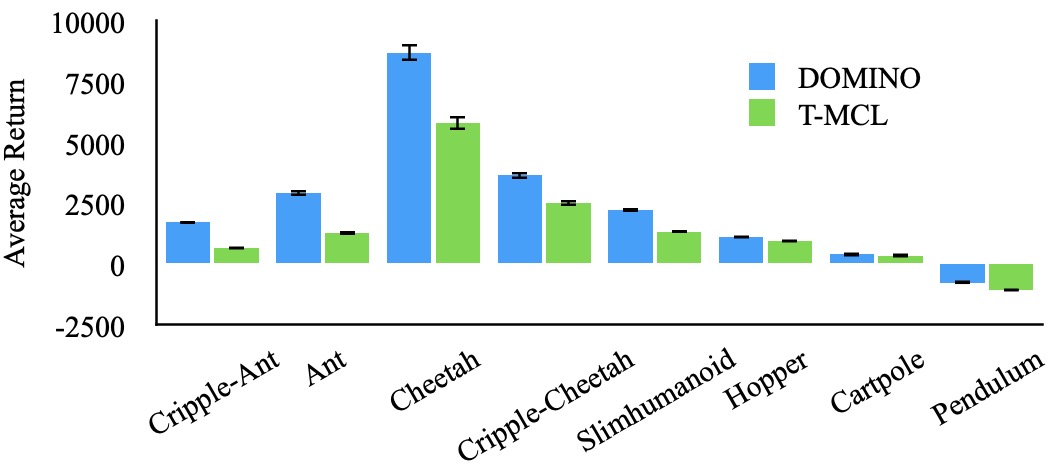}
         \caption{ {Performance comparison}}
         \label{fig:Performance Comparison}
     \end{subfigure}
        \caption{Generalization with complex dynamics changes. The transition dynamics of the robot may simultaneously influenced by multiple confounders, such as mass (\imgintext{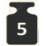}), length of leg (\imgintext{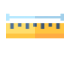}),  or a crippled leg (\imgintext{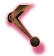}).  In real-world situations, all the possible confounders may change simultaneously, which bring challenges to robotic dynamics generalization. 
        DOMINO addresses such problem by decomposed MI optimization and achieves the state-of-the-art performance. 
        }
        \vspace{-10pt}
\end{figure}

Recent advanced context-aware meta-RL methods \cite{fu2020towards, wang2021improving,li2021provably,sang2022pandr} further improve meta-RL via contrastive learning, which optimizes the InfoNCE bound \cite{oord2018representation} of the mutual information in essence. These methods show a promising improvement in entangled context learning, which performs well in single confounded environments.
However, as demonstrated in Figure \ref{fig:Dynamic Generalization},
in real-world situations for robotic applications with partially unspecified dynamics, the transition dynamics can be influenced by multiple confounders simultaneously, such as mass changes, damping, friction, or  malfunctional modules like a crippled leg. 
For example, when a transportation robot is working in the wild, the load will dynamically change as the task progresses, while the humidity and roughness of the road also may vary. 
 {
Moreover, some works also construct a confounder set for unsupervised RL environment generalization\cite{dennis2020emergent, jiang2021prioritized, jiang2021replay, parker2022evolving, raileanu2021decoupling, cobbe2019leveraging}}.   {RIA \cite{guo2021relational} also constructs confounder sets with multiple confounders for unsupervised dynamics generalization.} Such changeable environments bring great challenges to the robot for capturing contextual information, which motivates our study. 

\textbf{Contribution. } 
In this paper, we give a theoretical analysis which 
demonstrates that when the number of confounders increases, InfoNCE will be a loose bound of mutual information (MI) with the samples in limited seen environments, which is called MI underestimation~\cite{sordoni2021decomposed}. To tackle this 
problem, we  propose a \textbf{D}ec\textbf{O}mposed \textbf{M}utual \textbf{IN}formation \textbf{O}ptimization (\modelname) framework for context learning in meta-RL.  {The context encoder aims to embed the past state-action pairs into disentangled context vectors and is optimized by maximizing the mutual information between the disentangled context vectors and historical trajectories while minimizing the state transition prediction error.} \modelname decomposes the full MI optimization problem into a summation of $N$ smaller MI optimization problems by learning disentangled context. We then theoretically prove that DOMINO could alleviate the underestimation bias of the InfoNCE and reduce the demand for the samples collected in various environments \cite{mcallester2020formal,stratos2018mutual}. 
Last, with the learned disentangled context, we further develop the context-aware model-based and model-free algorithms to learn the context-conditioned policy and illustrate that \modelname can consistently improve generalization performance in both ways to overcome the challenge of multi-confounded dynamics.

Extensive experiments demonstrate that DOMINO benefits meta-RL on both the generalization performance in unseen environments and sample efficiency during the training process under the challenging multi-confounded setting.
For example, as show in Figure \ref{fig:Performance Comparison}, it achieves 1.5 times performance improvement to T-MCL \cite{lee2020context} in the Cheetah domain and 2.6 times performance improvement to T-MCL in the Crippled-Ant domain.
Visualization of the learned context demonstrates that the disentangled context generated by DOMINO under different environments could be more clearly distinguished in the embedding space, which indicates its advantage to extract high-quality contextual information from the environment.

\section{Related Work} \label{sec:related_work}
\subsection{Meta-Reinforcement Learning}
Meta-RL extends the framework of meta-learning~\cite{Schmidhuber1987EvolutionaryPI,Thrun1998LearningTL} to reinforcement learning, aiming to learn
an adaptive policy being able to generalize to unseen tasks. Specifically, meta-RL methods learn the policy based on the prior knowledge discovered from various training environments and reuse the policy to fast adapt to unseen testing environments after zero or few shots. Gradient-based meta-RL algorithms ~\cite{DBLP:conf/icml/FinnAL17,DBLP:conf/iclr/RothfussLCAA19,Liu2019TamingME,Gupta2018MetaReinforcementLO} learn a model initialization and adapt the parameters with few policy gradient updates in new dynamics. 
Context-based meta-RL algorithms \cite{DBLP:conf/icml/RakellyZFLQ19, DBLP:conf/iclr/FakoorCSS20,lee2020context,seo2020trajectory} learn contextual information to capture local dynamics explicitly and show great potential to tackle generalization tasks in complicated environments. Many model-free context-based methods are proposed to learn a policy conditioned on the latent context that can adapt with off-policy data by leveraging context information and is trained by maximizing the expected return.  PEARL~\cite{DBLP:conf/icml/RakellyZFLQ19} adapts to a new environment by inferring latent context variables from a small number of trajectories. Recent advanced methods further improve the quality of contextual representation leveraging contrastive learning \cite{fu2020towards, wang2021improving,li2021provably,sang2022pandr}.
Unlike the model-free methods mentioned above, context-aware world models are proposed to learn the dynamics with confounders directly. 
CaDM \cite{DBLP:journals/corr/abs-2005-06800} learns a global model that generalizes across tasks by training a latent context to capture the local dynamics. T-MCL \cite{seo2020trajectory}  combines multiple-choice learning with context-aware world model and achieves state-of-the-art results on the dynamics generalization tasks.     {RIA \cite{guo2021relational} further expands this method into unsupervised setting without environment label by intervention, and enhances the context learning via MI optimization.}

However, existing context-based approaches focus on learning entangled context, in which each trajectory is encoded into only one context vector. In a multi-confounding environment, learning entangled contexts requires orders of magnitude higher samples to capture accurate dynamics information. 
    {To tackle this challenge, different from RIA \cite{guo2021relational} and T-MCL \cite{seo2020trajectory} , \modelname infers several disentangled context vectors from a single trajectory and divides the whole MI optimization into the summation of smaller ones}. The proposed decomposed MI optimization reduces the amount of demand for diverse samples and thus improves the generalization of the policy to overcome the adaptation problem in multi-confounded unseen environments.

\subsection{Mutual Information Optimization for Representation Learning}
Representation learning based on mutual information (MI) maximization has been applied in various tasks such as computer vision~\cite{grill2020bootstrap,caron2020unsupervised}, natural language processing~\cite{mikolov2013efficient,stratos2018mutual}, and RL~\cite{mazoure2020deep}, exploiting noise-contrastive estimation (NCE)~\cite{gutmann2012noise}, InfoNCE~\cite{oord2018representation} and variational objectives ~\cite{hjelm2018learning}. InfoNCE has gained recent interest with respect to variational approaches due to its lower variance~\cite{song2019understanding} and superior performance in downstream tasks. However, InfoNCE may underestimate the true MI, given that it is limited by the number of samples. To tackle this problem, DEMI~\cite{sordoni2021decomposed} first scaffolds the total MI estimation into a sequence of smaller estimation problems. In this paper, since the confounders in the real world are commonly independent, we simplify the complexity of mutual information decomposition and eliminate the need to learn conditional mutual information as a sub-term, assuming that multiple confounders are independent of each other.  
\section{Preliminaries}
\label{sec:problem_setup}

We consider standard RL framework where an agent optimizes a specified reward function through interacting with an environment.
Formally, we formulate our problem as a Markov decision process (MDP) \cite{sutton2018reinforcement}, which is defined as a tuple $\left( \mathcal{S}, \mathcal{A}, p, r, \gamma, \rho_0\right)$.
Here, $\mathcal{S}$ is the state space, 
$\mathcal{A}$ is the action space, 
$p\left(s^\prime| s,a\right)$ is the transition dynamics,
$r\left(s,a\right)$ is the reward function,
$\rho_0$ is the initial state distribution,
and $\gamma \in [0,1)$ is the discount factor.
In order to address the problem of generalization, we further consider the distribution of MDPs, where the transition dynamics $p_{\tilde{u}}\left(s^\prime| s,a \right)$ varies according to multiple confounders  $\tilde{u} = \left\{u_{0},u_{1},\ldots,u_{N}\right\}$. The confounders can be continuous random variables, like the mass, damping, random disturbance force, or discrete random variables, such as one of the robot's leg is crippled.
We assume that the true transition dynamics model is unknown, but the state transition data can be sampled by taking actions in the environment. Given a set of training setting sampled from $p(\tilde{u}_{\text{train}})$, the meta-training process learns a policy $\pi(s,c)$ that adapts to the task at hand by conditioning on the embedding of the history of past transitions, which we refer as context $c$. At test-time, the policy should adapt to the new MDP under the test setting $\tilde{u}_{\text{test}}$ drawn from $p(\tilde{u}_{\text{test}})$.   

{
Our goal is to learn a policy to maximizing the expected return $\mathcal{R}_{\text{train}}$ condition on the context $c$ which is encoded from the sequences of current state action pairs $\{s_{\tau},a_{\tau},s_{\tau+1}\}_{\tau=t-H}^{t}$ in several training scenarios and enable it to perform well and achieve a high expected return $\mathcal{R}_{test}$ in test scenarios never seen before.}



\begin{equation}
\footnotesize
{
\max_{\pi} \left\{\mathcal{R}_{\#} = \mathbb{E}_{\tilde{u} \sim p(\tilde{u}_{\#})}\left[\sum_{t=0}^{\infty} \gamma^{t} r\left(\mathbf{s}_{t},  \mathbf{a}_{t}\right)\right]\right\},\quad  a_{t} \sim \pi(s_{t},c), \quad \#=\left\{\text{"train" or "test"}\right\}}
\end{equation}

 \begin{figure}[t]
 \label{fig:framework}
    \centering
    \includegraphics[width=0.95\textwidth]{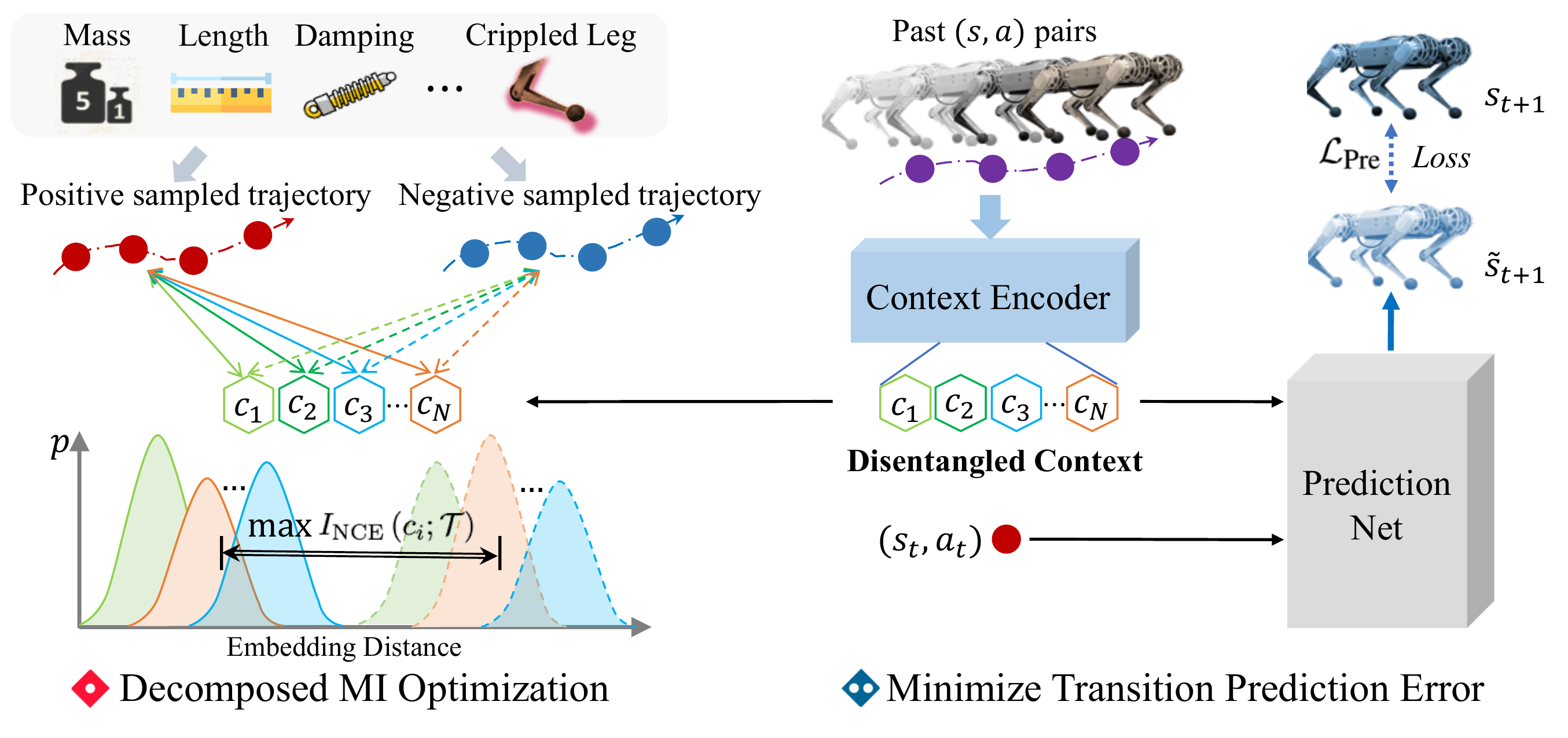}
    \caption{The overall framework of \modelname (\imgintext{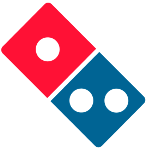}). The context encoder embeds the past state-action pairs (\imgintext{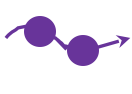})  into disentangled context vectors.   The disentangled context vectors (\imgintext{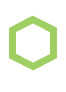}) are learned via the decomposed mutual information optimization (\imgintext{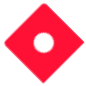}) while minimizing the state transition prediction error (\imgintext{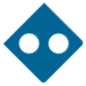}).
    \textbf{Minimize transition prediction error}: With the current state-action pair and the learned context vectors, future state can be predicted by the prediction network. The gradient of the prediction error will be used to update both the context encoder and the prediction network.
    \textbf{Decomposed MI Optimization}: we optimize the MI between the learned context and the historical trajectories under the same confounder setting via maximizing the InfoNCE bound, which aims to minimize the embedding distance between the positive sampled trajectories (\imgintext{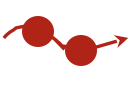}) and the context, while maximizing the embedding distance between the negative sampled trajectories (\imgintext{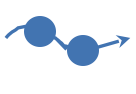}) and the context. 
 }
    \vspace{-10pt}
    
\end{figure}
\section{Decomposed Mutual Information Optimization for Context Learning}


In this section, we first provide a theoretical analysis to show why the multi-confounded environments are more challenging. We find that when the number of confounders increases, the InfoNCE will be a loose bound of MI with the samples in limited seen environments, resulting in the underestimation of MI.   To solve such a problem, we develop the \modelname framework to learn disentangled context by decomposed MI optimization. We theoretically illustrate that the decomposed MI optimization can alleviate the underestimation of MI and reduce the demand for the number of samples. 
The disentangled context $c=\left\{c_{0},c_{1},\ldots,c_{N}\right\}$ is embedded by the context encoder with parameter $\varphi$ from the past state-action pairs $\tau^{*}=\left\{s^{*}_{l},a^{*}_{l}\right\}_{l=t-H}^{t-1}$ in current episode.
\modelname explicitly maximizes the MI between the context $c$ and the historical trajectories $\mathcal{T}=\left\{\tau^{i}\right\}_{i=1}^{M}$ ($\tau^{i}=\left\{s^{i}_{l},a^{i}_{l}\right\}_{l=0}^{T}$) collected based on the combination of multiple confounders $\tilde{u}=\left\{u_{0},u_{1},\dots,u_{N}\right\}$ as same as the current confounder setting, while minimizing the state transition prediction error conditioned on the learned context. 
We solve the MI optimization problem by optimizing the InfoNCE lower bound on MI~\cite{oord2018representation}, which can be viewed as a contrastive method for the MI optimization, and decompose the full MI optimization into smaller ones to alleviate the underestimation of the mutual information and reduce the demand for the number of samples collected in various environments.



\subsection{InfoNCE Bound for  Mutual Information Optimization}
InfoNCE bound $I_{\text{NCE}}(x; y)$ is a lower bound of the mutual information $I(x; y)$, where NCE stands for Noise-Contrastive Estimation, is a type of contrastive loss function used for self-supervised learning.  InfoNCE is obtained by comparing pairs sampled from the joint distribution $x, y_1 \sim p(x, y)$ ($y_1$ is called the positive example) to pairs $x, y_i$ built using a set of negative examples, $y_{2:K} \sim p(y_{2:K}) = \prod_{k = 2}^K p(y_k)$:
\begin{equation}
\footnotesize
I(x; y) \geq I_{\text{NCE}}(x; y \mid \psi, K)= E\left[\log \frac{e^{\psi\left(x, y_{1}\right)}}{\frac{1}{K} \sum_{k=1}^{K} e^{\psi\left(x, y_{k}\right)}}\right] 
\label{eq:infonce}
\end{equation}
where  $\psi$ is a  function assigning a similarity score to $x, y$ pairs and $K$ denotes the number of  samples. Through discriminating naturally the paired positive instances from the randomly paired negative instances, it is proved to bring universal performance gains in various domains, such as computer vision and natural language processing.

\begin{lemma}
\label{info-lemma}
$I_{\mathrm{NCE}}(X; Y\mid K) \leq I(x; y) \leq \log K$ \textit{is a necessary condition for} $I_{\mathrm{NCE}}(X;Y \mid K)$ \textit{to be a tight bound of $I(x; y)$.} (see proof in Appendix  \ref{app:proof_nce})
\end{lemma}


Some previous context-aware methods 
learn an entangled context $c$ by maximizing the mutual information between the context $c$ embedded from the past state-action pairs in the current episode, and the historical trajectories $\mathcal{T}$ collected under the same confounder setting as the current episode. 
They solve this problem by maximizing the InfoNCE lower bound on $I_{\text{NCE}}(c;\mathcal{T})$, which can be viewed as a contrastive estimation ~\cite{oord2018representation} of $I_{\text{NCE}}(c;\mathcal{T})$, and obtain promising improvement in single-confounded environment.
 However, according to Lemma \ref{info-lemma}, the $I_{\text{NCE}}( c;\mathcal{T})$ may be loose if the true mutual information $I(c; \mathcal{T})$ is larger than $\log K$, which is called underestimation of the mutual information. Therefore, to make the InfoNCE bound to be a tight bound of $I(c;\mathcal{T})$, the minimum number of samples is $e^{I(c;\mathcal{T})}$.
In real-world robotic control tasks, the dynamics of the robot is commonly influenced by multiple confounders $\tilde{u}=\left\{u_{0},\ldots,u_{i},\ldots,u_{j},\ldots,u_{N}\right\}$ simultaneously, under the assumption that the confounders are independent (such as mass and damping), the mutual information  between the historical trajectories $\mathcal{T}$ and the context $c$  can be derived as 
\begin{equation}
\footnotesize
\begin{split}
I(c;\mathcal{T}) & =\mathbb{E}_{p(\tau, c)} \left\{\log \frac{p(\mathcal{T} \mid c)}{p(\mathcal{T})}\right\}=\mathbb{E}_{p(\tau, c)} \log \left\{\frac{\int p(\mathcal{T} \mid \tilde{u}) p(\tilde{u} \mid c) \mathrm{d} \tilde{u}}{p(\mathcal{T})}\right\}\\
&\geq \mathbb{E}_{p(\tau, c)p(\tilde{u} \mid c)} \left\{ \log \frac{p(\mathcal{T} \mid \tilde{u})}{p(\mathcal{T})}\right\}=I( \tilde{u};\mathcal{T})\stackrel{u_{i} \perp u_{j} }{\Longrightarrow}\sum_{i=0}^{N} I\left(u_{i};\mathcal{T} \right)
\end{split}
\end{equation}
{As the number of confounders increases, the lower bound of $I(c; \mathcal{T})$ will become larger, and the necessary condition for $I_{\text{NCE}}(\mathcal{T};c \mid K)$ to be a tight bound of $I(c; \mathcal{T})$ will become more difficult to satisfy. Since $I(c;\mathcal{T}) \geq \sum_{i=0}^{N} I\left( u_{i};\mathcal{T} \right)$, to let the necessary consition satisfied, the amount of data $K$ must be larger than $e^{\sum_{i=0}^{N} I\left( u_{i};\mathcal{T} \right)}$ according to Lemma \ref{info-lemma}. Thus the demand for data increases significantly.} Since the confounders are commonly independent in real-world, can we relax this condition by learning disentangled context vectors instead of entangled context intuitively?



\subsection{Decomposed MI Optimization}
\label{theoretical results}




If the context vectors $c=\{c_{0},c_{1},\ldots,c_{N}\}$ can be independent, then we can ease this problem by applying the chain rule on MI to decompose the total MI into a sum of  smaller MI terms, i.e.,
\begin{equation}
\footnotesize
I_{\text{NCE}}( c;\mathcal{T} \mid K) = \sum_{i=0}^{N} \left\{I_{\text{NCE}}(c_{i};\mathcal{T} \mid K)\right\} \leq N\log K
\end{equation}
\begin{theorem}
\label{info-thm}
\vspace{-10pt}
If the context vectors $\left\{c_{0},c_{1},\ldots,c_{N}\right\}$  can be independent, then the necessary condition for $I_{\mathrm{NCE}}(c;\mathcal{T})$ to be a tight bound can be relaxed to $I(c;\mathcal{T}) \leq N\log K = \log K^{N}$. 
Thus, the need of the number of  samples can be reduced from $K \geq e^{I(c;\mathcal{T})}$ to $K \geq e^{\frac{1}{N}I(c;\mathcal{T})}$. 
\end{theorem}

Inspired by Theorem \ref{info-thm}, we intuitively  learn disentangled context vectors and maximize the mutual information between the historical trajectories $\mathcal{T}$ and the context vectors $\left\{c_{0},\ldots,c_{N}\right\}$ while minimizing the $I_{\text{NCE}}$ between the context vectors,  
i.e., to maximize the $\mathcal{L}_{\text{NCE}}$
\begin{equation}
\footnotesize
{
\mathcal{L}_{\text{NCE}}(\varphi,w)= \sum_{i=0}^N I_{\text{NCE}}(c_{i};\mathcal{T})- \sum_{j=0}^{N}\sum_{i=0,i \neq j}^{N} I_{\text{NCE}}(c_{i};c_{j})}
\label{eq:L_nce}
\end{equation}


where the $I_{\text{NCE}}(c_{i};\mathcal{T})$ can be obtained with the positive trajectory $\tau^{+}$ and negative trajectories $\left\{\tau^{-}_{k}\right\}_{k=2}^{K}$, i.e.,
\vspace{-10pt}
\begin{equation}
\footnotesize
I_{\mathrm{NCE}}\left(c_{i} ; \mathcal{T}\right)=E\left[\log \frac{e^{\psi\left(c_{i}, h_{w}\left(\tau^{+}\right)\right)}}{\frac{1}{K}\left(\sum_{k=2}^{K} e^{\psi\left(c_{i}, h_{w}\left(\tau_{k}^{-}\right)\right)}+e^{\psi\left(c_{i}, h_{w}\left(\tau^{+}\right)\right)}\right)}\right]
\label{eq:IC_nce}
\end{equation}

The context $c=\{c_{0},\ldots,c_{i},\ldots,c_{N}\}$  ($c_{i}\in \mathbb{R}^{m}$) is encoded from the past state action pairs $\tau^{*}$ in current episode by $g_{\varphi}(\cdot)$. Both the positive trajectory (collected in same setting of the confounders) and  negative trajectory (collected in different setting of the confounders) are encoded   to $\mathbb{R}^{m}$ by $h_{w}(\cdot)$. {
The critic function $\psi(\cdot,\cdot)$   measures the cosine similarity between inputs by dot product after normalization.}
Under the assumption that the setting of confounders will not change in one episode, to obtain the $I_{\text{NCE}}(c_{i};c_{j})$, we use $c^{+}_{j}$ sampled from same episode like $c_{i}$ as the positive example, and use $c^{-}_{j}$ sampled from different episode as the negative example. 
Thus the $I_{\text{NCE}}(c_{i};c_{j})$ can be derived as 
\begin{equation}
\footnotesize
I_{\mathrm{NCE}}\left(c_{i} ; c_{j}\right)=E\left[\log \frac{e^{\psi\left(c_{i}, c_{j}^{+}\right)}}{\frac{1}{K}\left(\sum_{k=2}^{K} e^{\psi\left(c_{i}, c_{j}^{k-}\right)}+e^{\psi\left(c_{i}, c_{j}^{+}\right)}\right)}\right]
\label{eq:IC_nce}
\end{equation}

Then, the future state $s_{t+1}$  can be predicted with the the current state $s_{t}$, action $a_{t}$ and the disentangled context vectors $\left\{c_{0_{t}},\ldots,c_{N_{t}}\right\}$ by the state transition prediction network $f_{\phi}(\cdot)$. We aim to minimize the prediction loss, which is equal to maximizing
\begin{equation}
\footnotesize
{
 \mathcal{L}_{\text{Pre}}(\varphi,\phi)=E_{\tau^{*} \sim \mathcal{B}}\left[-\frac{1}{H} \sum_{\lambda=t}^{t+H-1} \log f_{\phi}\left(s_{i+1} \mid s_{i}, a_{i},\left(c_{{0}_{\lambda}}, \ldots, c_{{N}_{\lambda}}\right)\right)\right],\tau^{*}=\left\{s^{*}_{l},a^{*}_{l}\right\}_{l=t-H}^{t-1}  }
\end{equation}
where $\mathcal{B}$ is the training set, and $H$ is the prediction horizon.
The whole framework of \modelname is demonstrated in Figure \ref{fig:framework} and the overall objective function of \modelname is 
\begin{equation}
{
 \mathcal{L}(\varphi,w,\phi)=\mathcal{L}_{\text{Pre}}(\varphi,\phi)+\mathcal{L}_{\text{NCE}}(\varphi,w)}
\end{equation}

\subsection{Combine \modelname with Downstream RL Methods}
\textbf{Combination with Model-based RL. }
With DOMINO we can learn the context encoder and the context-aware world model together. First, the past state-action pairs are encoded into the disentangled context vectors by the context encoder. According to the learned context, the transition prediction network predicts the future states of different actions. 
In particular, we use the cross entropy method (CEM) \cite{botev2013cross}, a typical neural model predictive control (MPC) \cite{garcia1989model} method, to select actions, in which several candidate action sequences are iteratively sampled from a candidate distribution, which is adjusted based on best-performing action samples. The optimal action sequence $\boldsymbol{a}_{t: t+T} \doteq\left\{\boldsymbol{a}_{t}, \ldots, \boldsymbol{a}_{t+T}\right\}$ can be obtained by 
\begin{equation}
\operatorname{argmax}_{a_{t: t+T}} \sum_{\lambda=t}^{t+T} \mathbb{E}_{\tilde{f}}\left[r\left(\boldsymbol{s}_{\lambda}, \boldsymbol{a}_{\lambda}\right)\right], \quad \tilde{f}=\operatorname{Pr}\left(s_{t+1} \mid s_{t}, a_{t},c_{t_{0}},\ldots,c_{t_{N}}\right)
\end{equation}
Then, we use the mean value of adjusted candidate distribution as action and re-plan at every timestep. We provide detailed algorithm pseudo-code in the Appendix \ref{model-based pseudo-code}.
{
As for the adaptation process, the policy and context encoder zero-shot adapts to the unseen confounders setting $u_{test}$, and we use the same adaptive planning method as T-CML\cite{lee2020context}, which selects the most accurate prediction head over a recent experience condition on the inferred context. The details is introduced in Appendix \ref{adaptive planning}}

\textbf{Combination with Model-free RL. }
 Previous works show that a policy learned by model-free method can be more robust to dynamics changes when it takes the contextual information as an additional input \cite{yu2017preparing,packer2018assessing,zhou2019environment}. Motivated by this, we investigate whether the context encoder learned by \modelname can be used as a plug-and-play module to improve the final generalization performance of model-free RL methods. We concatenate the disentangled context encoded by a pre-trained context encoder from \modelname and the current state-action pairs, and learn a conditional policy $\pi \left(a_t|s_t,c_{0},\ldots,c_{N}\right)$.
We use the Proximal Policy Optimization (PPO) method to train the agent \cite{schulman2017proximal}, which learns the policy by maximizing
\begin{equation}
\footnotesize
\hat{\mathbb{E}}_{t}\left[\frac{\pi\left(a_{t} \mid s_{t},c_{t_{0}},\ldots,c_{t_{N}}\right)}{\pi_{\theta_{\text {old }}}\left(a_{t} \mid s_{t},c_{t_{0}},\ldots,c_{t_{N}}\right)} \hat{A}_{t}-\beta \mathrm{KL}\left[\pi_{\theta_{\text {old }}}\left(\cdot \mid s_{t},c_{t_{0}},\ldots,c_{t_{N}}\right), \pi \left(\cdot \mid s_{t},c_{0},\ldots,c_{N}\right)\right]\right]
\end{equation}
where $\hat{A}_{t}$ is the estimation of the advantage function at timestep $t$. We provide detailed pseudo-code in the Appendix \ref{model-free pseudo-code}.

\section{Experiments} \label{sec:exp}
\begin{figure}[t]
    \centering
    \includegraphics[width=0.95\linewidth]{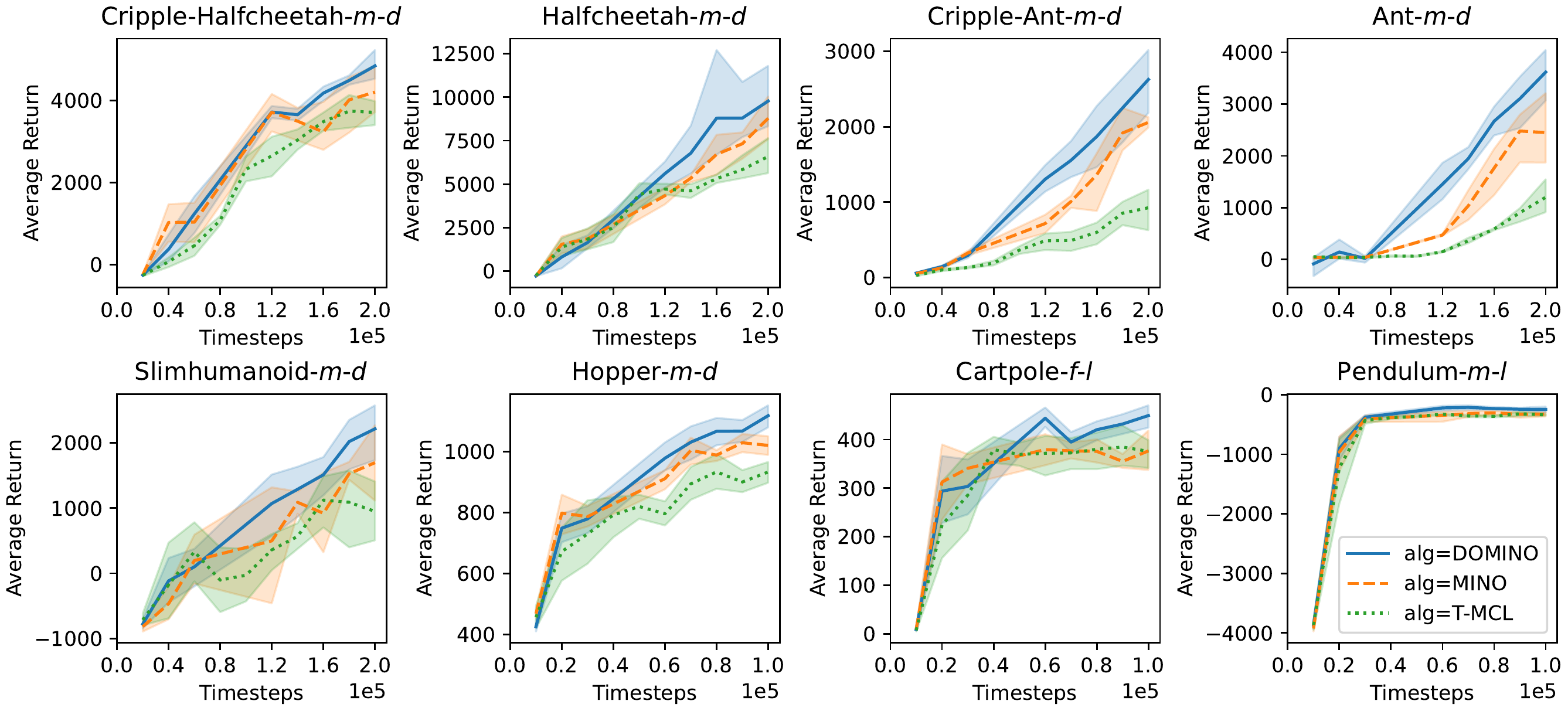}
    \caption{The average returns of the model-based methods in training  environments (over 8 seeds).}
    \label{fig:mb_demic_train}
\end{figure}

In this section, we evaluate the performance of our \modelname method to answer the following questions:
\textbf{(1)} Can \modelname help the model-based RL methods overcome the   multi-confounded challenges in dynamics generalization (see comparison with ablation in Figure \ref{fig:mb_demic_train} and Figure \ref{fig:mb_demic_test})? \textbf{(2)} Can the context encoder learned by \modelname be used as a plug-and-play module to improve the generalization abilities of model-free RL methods in multi-confounded environments (see comparison with ablation in Table \ref{tab:mf-compare} and Table \ref{tab:mf-compare2})? \textbf{(3)}
Can the proposed decomposed MI optimization benefit the forward prediction of the world model? (see Figure \ref{fig:pre_arc})
\textbf{(4)} Does the disentangled context extract more meaningful contextual information than entangled context (see Figure \ref{fig:vis graphs})?

\subsection{Setups}
\label{sec-Setups}
We demonstrate the effectiveness of our proposed method on 8 benchmarks, which contain 6  typical robotic control tasks based on the MuJoCo physics engine \cite{todorov2012mujoco}  and 2 classical control tasks (CartPole and Pendulum) from OpenAI Gym \cite{brockman2016openai}. Different from previous works, all the environments are influenced by multiple confounders simultaneously. In our experiments, we modify multiple environment parameters at the same time (e.g., mass, length, damping, push force, and crippled leg) that characterize the transition dynamics. The robotic control tasks contain 4 environments (Hopper, HalfCheetah, Ant, SlimHumanoid) affected by multiple continuous confounders and 2 more difficult environments (Crippled Ant and Crippled HalfCheetah) affected by both continuous and discrete confounders. 
The detailed settings are illustrated in Appendix \ref{Detail env} (Table \ref{tbl:environment}). We implement these environments based on the publicly available code provide by \cite{nagabandi2018learning,seo2020trajectory}, and we also open-source the code  of the multiple-confounded environments$\footnote{\scriptsize{https://anonymous.4open.science/r/Multiple-confounded-Mujoco-Envs-01F3}}$.
For both training and testing phase, we sample the confounders at the beginning of each episode. During training, we randomly 
select a combination of confounders from a training set. At test time, we evaluate each algorithm in unseen environments with confounders outside the training range. 

\subsection{Comparison with Model-based Methods}
\textbf{Baselines. }  {We consider T-MCL \cite{lee2020context} and RIA\cite{guo2021relational} as the key baselines in comparison with model-based methods, which achieve the state-of-the-art results in zero-shot dynamics generalization tasks. Since RIA doesn't has a adaptive planning process, we provide the DOMINO and T-MCL without adaptive planning to fair compare to the RIA.}
We also consider an \textit{\textbf{\textcolor{codegray2}{ablation version}}} of \modelname as a baseline (denoted as MINO) to show the effectiveness of the decomposed MI optimization, which optimizes the MI and predicts the future states with an entangled context without decomposition.

\begin{figure}[hbpt]

\begin{minipage}{0.53\textwidth}
    \textbf{Results. } 
 {As shown in Figure \ref{fig:ria-comp}, DOMINO achieves better generalization performance than RIA and TMCL even without the adaptive planning, especially in complex environments like  Halfcheetah-$m$-$d$ and Slim-humanoid-$m$-$d$.}
Figure \ref{fig:mb_demic_train} shows the average return during the learning process in the training environments. The results illustrate that \modelname learns the policy more efficiently  than T-MCL and MINO.  
\end{minipage}
\hfill
\begin{minipage}[htbp]{0.45\textwidth}
    \centering
    \includegraphics[width=\linewidth]{ 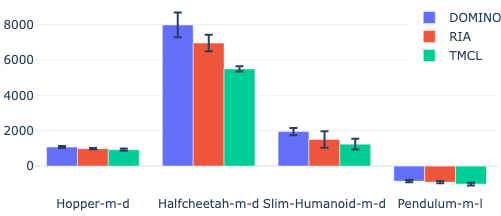}
\label{fig:ria-comp}
 \caption{ {Comparison w/o adaptive planning}}
\end{minipage}
\end{figure}

\begin{figure}[t]
    \centering
    \includegraphics[width=0.95\linewidth]{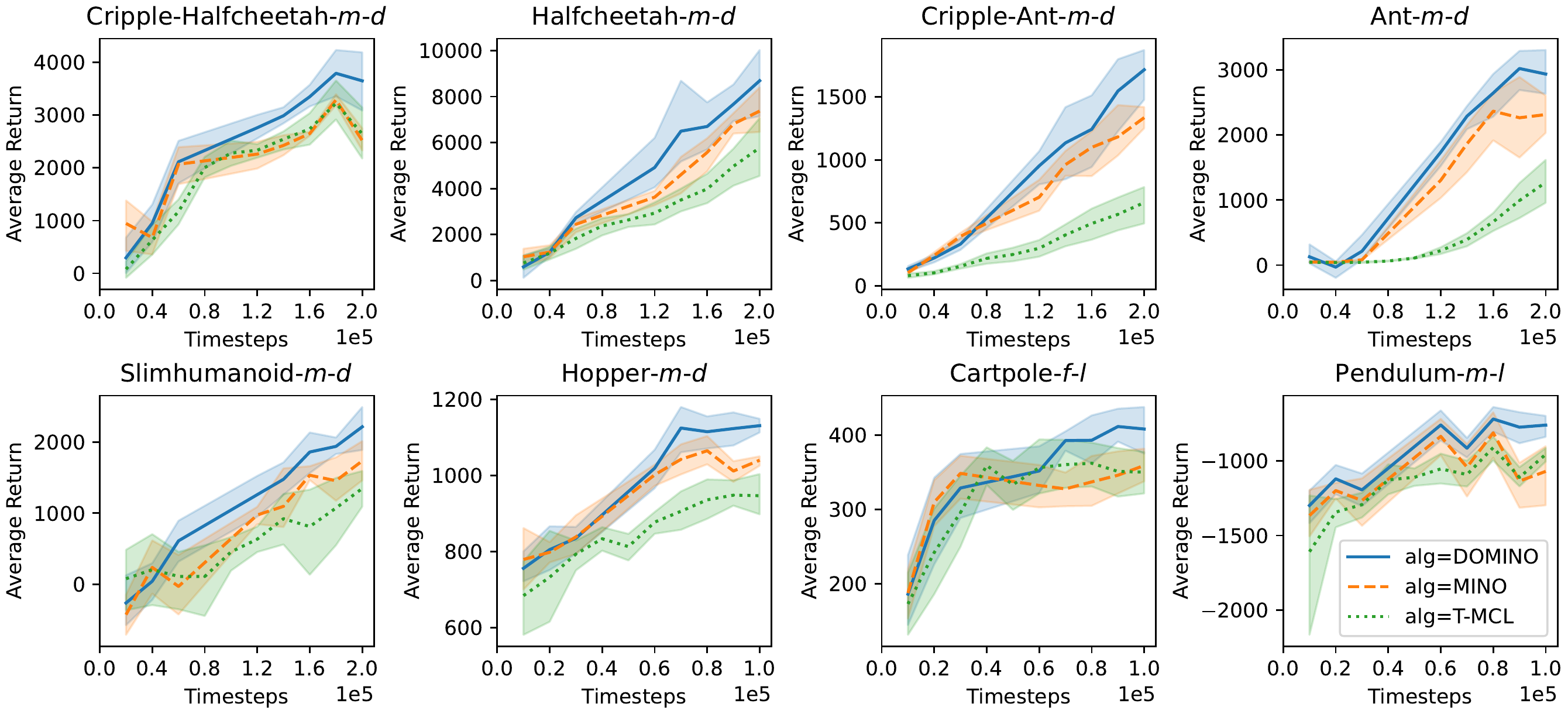}
    \caption{
Comparison with \textit{model-based methods} of the generalization performance (average return)  in unseen   multi-confounded environments (over 8 seeds).   }
    \label{fig:mb_demic_test}
\end{figure}

Figure \ref{fig:mb_demic_test}  shows the generalization performance tested in the unseen environments. 
 The results show that \modelname surpasses T-MCL in terms of the generalization performance and the learning sample efficiency. This demonstrates that the disentangled context improves the context-aware world model. Especially, the performance gain becomes much more significant in more complex environments (e.g., long-horizon and high-dimensional domains like Cripple-Ant, Ant, and Hopper). For example, \modelname achieves about 2.6 times  improvement to 
T-MCL in Cripple-Ant-$m$-$d$, which is one of the most difficult environment, whose leg will randomly be crippled, and its mass and damping will be changed in testing. 
More details are shown in Appendix \ref{app_detail}.

\subsection{Comparison with Model-free Methods}
We also verify whether the learned disentangled context is useful for improving the generalization performance of model-free RL methods. Similar to \cite{lee2020context,henderson2018deep}, we use the Proximal Policy Optimization  (PPO \cite{schulman2017proximal}) method to train the agents.

\textbf{Baselines}. Our proposed method, which takes the  context  learned by \modelname as conditional input  (PPO+\modelname), is compared with several context-conditional policies \cite{rakelly2019efficient, zhou2019environment}.
Specifically, we consider combining the PPO with the context learned by T-MCL (PPO+T-MCL), which learns the context encoder via a context-aware world model and achieves the state-of-the-art performance on dynamics generalization. We also consider PEARL \cite{rakelly2019efficient}, which learns probabilistic context variable by maximizing the expected returns.
We further develop an \textit{\textbf{\textcolor{codegray2}{ablation version}}} of \modelname, which optimize the MI with entangled context (PPO+MINO) as a baseline to illustrate the effectiveness of the decomposed MI optimization. 
We provide more detailed explanations in Appendix \ref{app_detail}. 

\textbf{Results}. Table \ref{tab:mf-compare} and Table \ref{tab:mf-compare2} show the performance of various model-free RL methods on both training and test environments.
PPO+\modelname shows superior performance and shows better generalization performances than previous conditional policy methods, implying that the proposed \modelname method can extract contextual information more effectively than both the context learned by the model-based method (PPO+T-MCL) and the context learned by the model-free method (PEARL).
Furthermore, PPO+\modelname shows an obvious advantage over PPO+MINO, especially in complex environments, such as HalfCheetah, Ant, and Hopper, which implies that the decomposed MI optimization improves the context learning significantly. 
Additionally, the results also show that compared to PEARL, the context learned by T-MCL and \modelname has better performance which  implies that the state transition perdition can help to extract contextual information more effectively. 

\begin{table}[htbp]
\caption{
Comparison with \textit{model-free} methods in the multi-confounded environments (over 5 seeds).  The transition dynamics will change in both training and test environments in every episode. }
\centering
\begin{adjustbox}{max width=0.9\textwidth}
\begin{tabular}{@{}lllllllll@{}}
\toprule
            & \multicolumn{2}{c}{Cartpole-$f$-$l$} &  & \multicolumn{2}{c}{Pendulum-$m$-$l$} &  & \multicolumn{2}{c}{Ant-$m$-$d$} \\ 
            \cline{2-3} \cline{5-6} \cline{8-9}
            & Train        & Test          &  & Train        & Test          &  & Train      & Test       \\
\toprule
PEARL       & 197$\pm{12}$       & 175$\pm{37}$        &  & -1265$\pm{173}$    & -1293$\pm{134}$   &  & 153$\pm{63}$     & 73$\pm{25}$     \\
PPO+T-MCL    & 220$\pm{27}$       & 182$\pm{25}$        &  & -558$\pm{184}$    & -579$\pm{128}$     &  & 176$\pm{82}$     & 173$\pm{38}$     \\
PPO+MINO & 267$\pm{18}$       & 234$\pm{46}$        &  & -497$\pm{162}$    & -526$\pm{219}$     &  & 194$\pm{95}$     & $184\pm{47}$     \\
\cellcolor{codegray}PPO+\modelname   & \cellcolor{codegray}\textbf{299}$\pm{23}$       & \cellcolor{codegray}\textbf{283}$\pm{68}$       & \cellcolor{codegray} & \cellcolor{codegray}\textbf{-405}$\pm{139}$    & \cellcolor{codegray}\textbf{-436}$\pm{146}$     & \cellcolor{codegray} & \cellcolor{codegray}\textbf{227}$\pm{86}$     & \cellcolor{codegray}\textbf{216}$\pm{52}$  \\ 
\bottomrule
 & \multicolumn{2}{c}{Halfcheetah-$m$-$d$} &  & \multicolumn{2}{c}{Slimhumanoid-$m$-$d$} &  & \multicolumn{2}{c}{Hopper-$m$-$d$} \\ 
            \cline{2-3} \cline{5-6} \cline{8-9}
            & Train        & Test          &  & Train        & Test          &  & Train      & Test       \\
\toprule
PEARL       & 1802$\pm{773}$       & 530$\pm{270}$        &  & 6947$\pm{3541}$    & 3697$\pm{2674}$   &  & 934$\pm{242}$     & 874$\pm{366}$     \\
PPO+T-MCL    & 2032$\pm{688}$       & 674$\pm{395}$        &  & 6157$\pm{1435}$    & 4136$\pm{1528}$     &  & 937$\pm{252}$     & 896$\pm{238}$     \\
PPO+MINO & 1973$\pm{563}$       & 824$\pm{498}$        &  & 6179$\pm{1123}$    & 4275$\pm{1134}$     &  & 1109$\pm{349}$     & 964$\pm{323}$     \\
\cellcolor{codegray}PPO+\modelname   & \cellcolor{codegray}\textbf{2472}$\pm{803}$       & \cellcolor{codegray}\textbf{1034}$\pm{476}$       & \cellcolor{codegray} & \cellcolor{codegray}\textbf{7825}$\pm{1256}$    & \cellcolor{codegray}\textbf{5258}$\pm{1039}$     &  \cellcolor{codegray}& \cellcolor{codegray}\textbf{1409}$\pm{254}$     & \cellcolor{codegray}\textbf{1137}$\pm{335}$  \\ \bottomrule
\end{tabular}
\label{tab:mf-compare}
\end{adjustbox}
\end{table}

\makeatletter
    \def\figcaption{%
        \refstepcounter{figure}%
        \@dblarg{\@caption{figure}}}
\makeatother

\begin{table}[t]
\vspace{-10pt}
\begin{minipage}{0.55\textwidth}
\caption{Comparison with \textit{model-free} methods in more difficult multi-confounded environments (over 5 seeds). }
\centering
\footnotesize
\begin{adjustbox}{max width=\textwidth}
\begin{tabular}{llllll}
\toprule
 & \multicolumn{2}{c}{Cripple-Ant-$m$-$d$} &  & \multicolumn{2}{c}{Cripple-Halfcheetah-$m$-$d$} \\ \hline
            & Train          & Test           &  & Train             & Test                \\
PEARL       & 182$\pm{73}$         & 96 $\pm{21}$       &  & \textbf{2538}$\pm{783}$         & 1028$\pm{445}$        \\
PPO+T-MCL    & 187$\pm{65}$         &109$\pm{36}$         &  & 2368$\pm{726}$        & 1006$\pm{434}$           \\
PPO+MINO & 206$\pm{64}$         & 113$\pm{34}$         &  & 2493$\pm{664}$         & 1197$\pm{424}$           \\
\cellcolor{codegray}PPO+\modelname   & \cellcolor{codegray}\textbf{233}$\pm{82}$         &\cellcolor{codegray}\textbf{132}$\pm{27}$        &\cellcolor{codegray}  &\cellcolor{codegray}2503$\pm{658}$        &\cellcolor{codegray}\textbf{1326}$\pm{491}$           \\ \bottomrule
\end{tabular}
\end{adjustbox}
\label{tab:mf-compare2}
\end{minipage}
\hfill
\begin{minipage}{0.44\textwidth}
    \vspace{15pt}
    \centering
    \includegraphics[width=\linewidth]{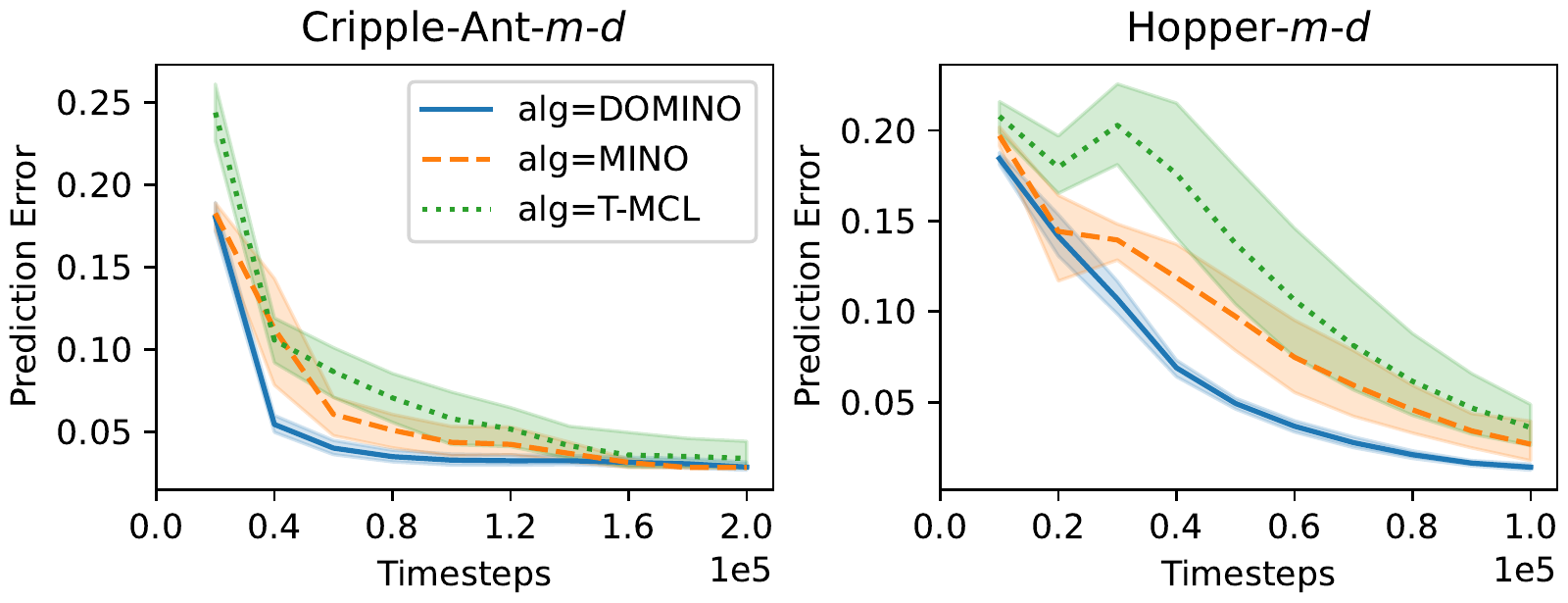}
    \vspace{-5pt}
    \figcaption{The testing prediction error.}
    \label{fig:pre_arc}
\end{minipage}
\vspace{-10pt}
\end{table}



\subsection{Disentangled Context Analysis}

\textbf{Prediction errors}. To show that our method indeed helps with a transition prediction, we compare baseline methods with DOMINO in terms of prediction error across 8 environments with varying multiple confounders. As shown in Figure \ref{fig:pre_arc}, our model demonstrates superior prediction performance, which indicates that  the learned context capture better contextual information compare with  entangled context (see more results in Appendix \ref{app-add}). 

\begin{figure}[htbp]

     \centering
     \begin{subfigure}[b]{0.45\textwidth}
         \centering
         \includegraphics[width=0.9\textwidth]{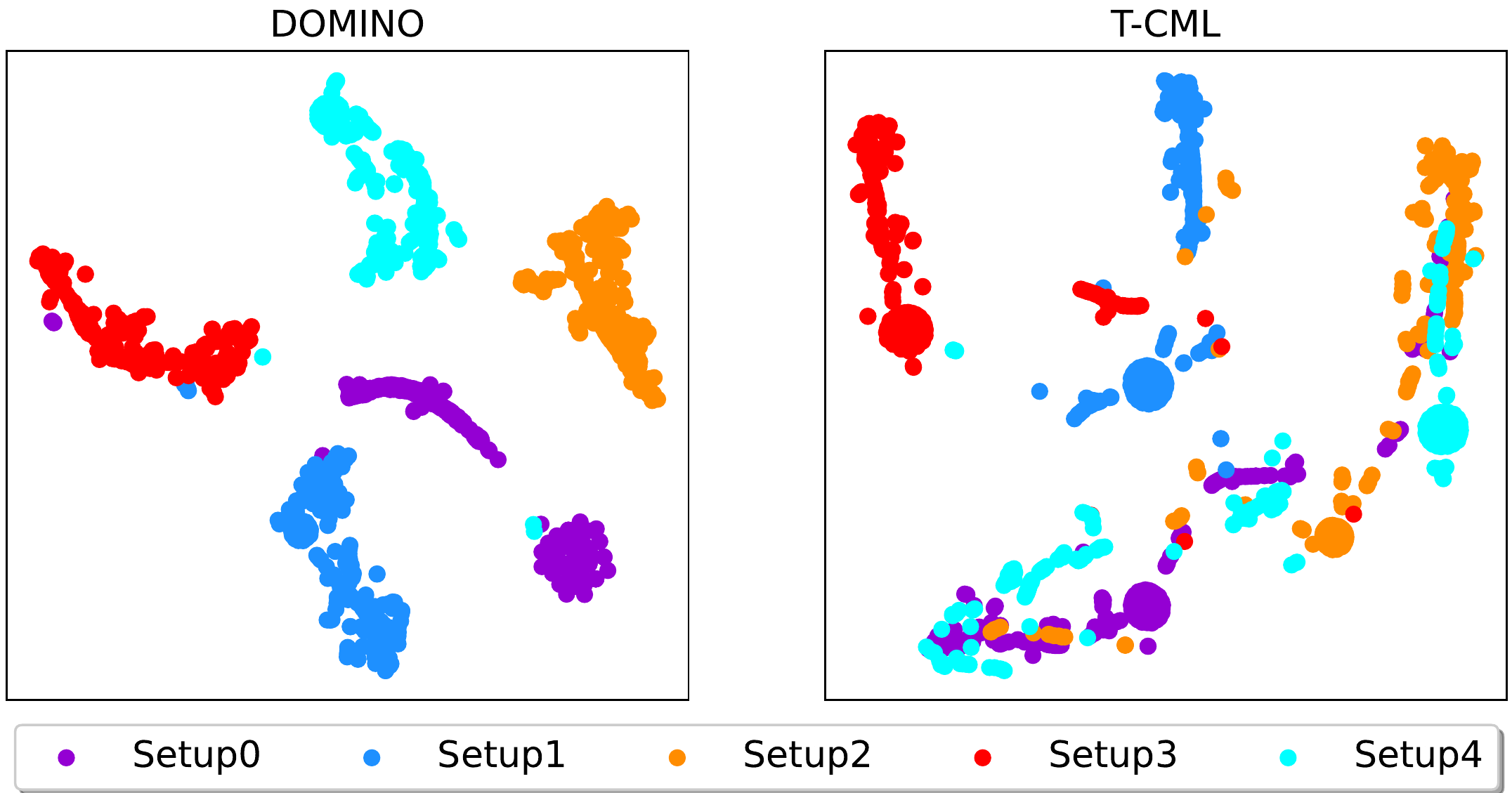}
         \caption{Visualization in Hopper-$m$-$d$.}
         
         \label{fig:y equals x}
     \end{subfigure}
     \hfill
     \begin{subfigure}[b]{0.45\textwidth}
         \centering
         \includegraphics[width=0.9\textwidth]{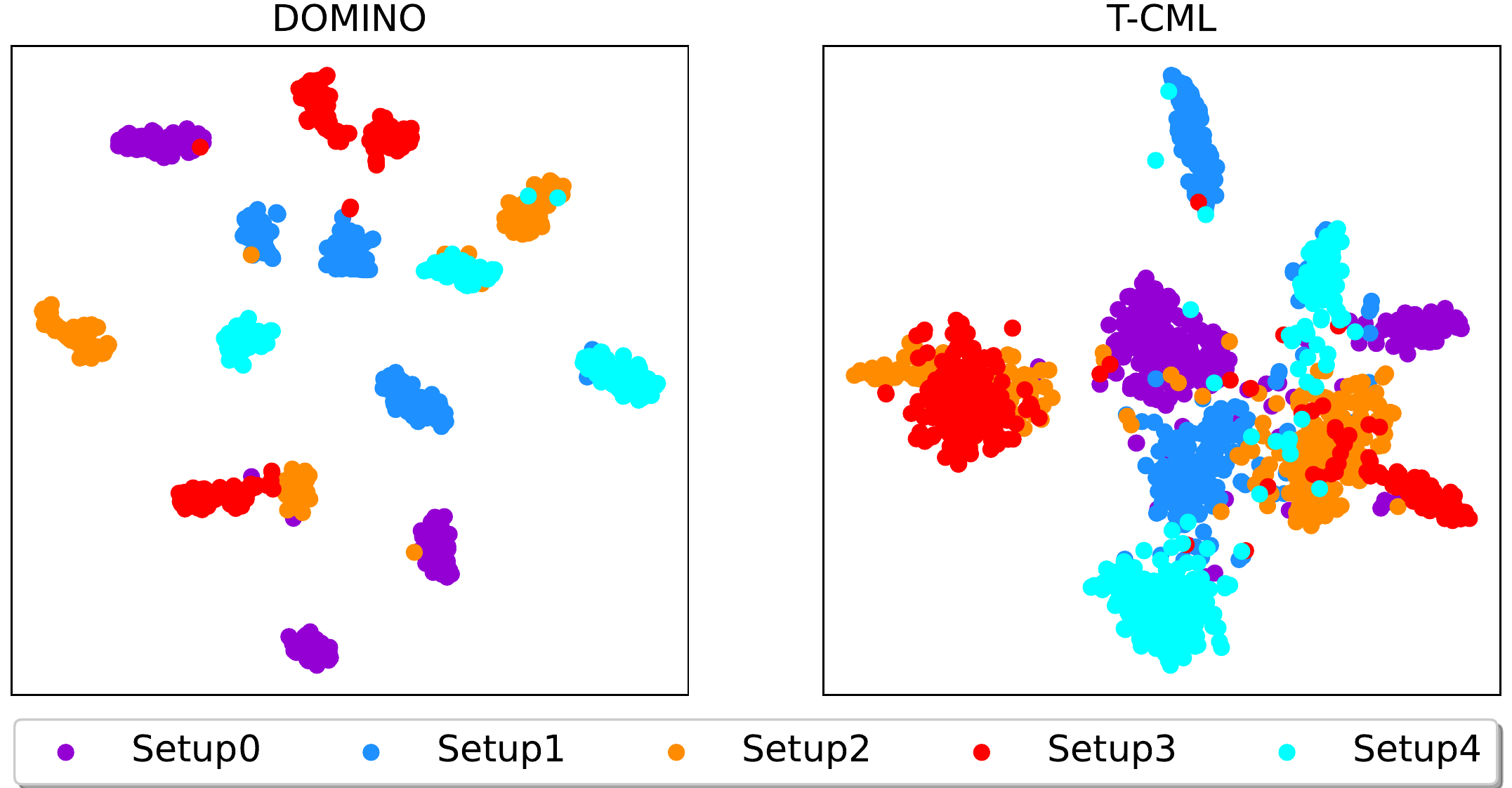}
         \caption{Visualization in Cripple-Ant-$m$.}
         
         \label{fig:three sin x}
     \end{subfigure}

        \caption{t-SNE visualization of context vectors extracted from trajectories collected in various environments. Embedded points from environments with the same confounders have the same color.
        \footnotesize{Hopper setups: \textit{Setup0} ($m=0.75, d=1.5$), \textit{Setup1} ($m=1.5, d=1.25$), \textit{Setup2} ($m=0.5, d=1.25$), \textit{Setup3} ($m=1.25, d=0.75$), \textit{Setup4} ($m=1.0, d=1.5$); Cripple-Ant setups: \textit{Setup0} ($m=1.15, leg=3$), \textit{Setup1} ($m=0.75, leg=1$), \textit{Setup2} ($m=1.25, leg=0$), \textit{Setup3} ($m=0.85, leg=0$), \textit{Setup4} ($m=1.0, leg=2$)(20 trajectories per setup)}.
        }
        \label{fig:vis graphs}

\end{figure}

\textbf{Visualization}. We visualize concatenation of the disentangled context vectors learned by \modelname via t-SNE \cite{maaten2008visualizing} and compare it with the entangled context learned by T-MCL.  As shown in Figure \ref{fig:vis graphs}, we find that the disentangled context vectors encoded from trajectories collected under different confounder settings 
could be more clearly distinguished in the embedding space than the entangled context learned by T-MCL.
This indicates that DOMINO extracts high-quality task-specific information from the environment compared with T-MCL.
We provide more visualization results  based on both t-SNE \cite{maaten2008visualizing}  and PCA \cite{jolliffe2002principal} in Appendix \ref{app-vis}.

\section{Conclusion}
In this paper, we propose a decomposed mutual information optimization (\modelname) framework to learn the generalized context for zero-shot dynamics generalization. The disentangled context is learned by maximizing the mutual information between the context and historical trajectories while minimizing the state transition prediction error. By decomposing the whole mutual information optimization problem into smaller ones, DOMINO can reduce the need for samples collected in various environments and overcome the underestimation of the mutual information in multi-confounded environments. Extensive experiments illustrate that DOMINO benefits the generalization performance in unseen environments with both model-based RL and model-free RL. {
For future work, an effective combination of DOMINO and RIA \cite{guo2021relational}, which expands the
decomposed MI optimization to  relational intervention approach proposed by RIA could become a stronger baseline for unsupervised dynamics generalization.} We believe our work can 
lay the foundation of dynamics generalization in complex environments. 

\textbf{Limitations and Negative Social Impact.}
DOMINO sets the number of disentangled context vectors as a hyper-parameter equal to the number of confounders in the environments, and capturing the number of confounders automatically  could be future works. 
We believe that DOMINO will not cause any negative social impact.

\section*{Acknowledgments and Disclosure of Funding}                   
The authors would like to thank the anonymous reviewers for their valuable comments and helpful suggestions. The work is supported by Huawei Noah’s Ark Lab; Ping Luo is supported by the General Research Fund of HK No.27208720, No.17212120, and No.17200622.

{\small
\bibliography{neurips_2022}
\bibliographystyle{unsrt}
}
\section*{Checklist}


\begin{enumerate}

\item For all authors...
\begin{enumerate}
  \item Do the main claims made in the abstract and introduction accurately reflect the paper's contributions and scope?
    \answerYes{}
  \item Did you describe the limitations of your work?
    \answerYes{}
  \item Did you discuss any potential negative societal impacts of your work?
    \answerYes{}
  \item Have you read the ethics review guidelines and ensured that your paper conforms to them?
    \answerYes{}
\end{enumerate}

\item If you are including theoretical results...
\begin{enumerate}
  \item Did you state the full set of assumptions of all theoretical results?
    \answerYes{}{See Section \ref{theoretical results}.}
        \item Did you include complete proofs of all theoretical results?
    \answerYes{}{See Section \ref{theoretical results} and Appendix \ref{app:proof_nce}.}
\end{enumerate}

\item If you ran experiments...
\begin{enumerate}
  \item Did you include the code, data, and instructions needed to reproduce the main experimental results (either in the supplemental material or as a URL)?
    \answerYes{}{See Section \ref{sec-Setups} and Appendix \ref{app_detail}.}
  \item Did you specify all the training details (e.g., data splits, hyperparameters, how they were chosen)?
    \answerYes{}{See Appendix \ref{app_detail}.}
        \item Did you report error bars (e.g., with respect to the random seed after running experiments multiple times)?
    \answerYes{}{See Section \ref{sec:exp}.}
        \item Did you include the total amount of compute and the type of resources used (e.g., type of GPUs, internal cluster, or cloud provider)?
    \answerYes{}{See Appendix \ref{app_detail}.}
\end{enumerate}

\item If you are using existing assets (e.g., code, data, models) or curating/releasing new assets...
\begin{enumerate}
  \item If your work uses existing assets, did you cite the creators?
    \answerYes{}
  \item Did you mention the license of the assets?
   \answerNA{}
  \item Did you include any new assets either in the supplemental material or as a URL?
    \answerYes{}
  \item Did you discuss whether and how consent was obtained from people whose data you're using/curating?
    \answerNA{}
  \item Did you discuss whether the data you are using/curating contains personally identifiable information or offensive content?
    \answerNA{}
\end{enumerate}

\item If you used crowdsourcing or conducted research with human subjects...
\begin{enumerate}
  \item Did you include the full text of instructions given to participants and screenshots, if applicable?
    \answerNA{}
  \item Did you describe any potential participant risks, with links to Institutional Review Board (IRB) approvals, if applicable?
    \answerNA{}
  \item Did you include the estimated hourly wage paid to participants and the total amount spent on participant compensation?
    \answerNA{}
\end{enumerate}

\end{enumerate}

\appendix
\newpage

\section{Derivations}

\label{app:proof_nce}
\subsection{  {Proof of Lemma 1}}
  {
According to the Barber and Agakov's variational lower bound~\cite{barber2003algorithm}, the mutual information  $\mi(\x; \xt)$ between $x$ and $y$ can be bounded as follows:}
\begin{equation}
  {
\mi(\x; \xt) = \expect_{p(\x, \xt)} \log \frac{p(\xt | \x)}{p(\xt)} \ge \expect_{p(\x, \xt)} \log \frac{q(\xt | \x)}{p(\xt)},}
\label{eq:barber-agakov-bound}
\end{equation}
  {
where $q$ is an arbitrary distribution.
Specifically, ${q(\xt | \x)}$ is defined by independently sampling a set of examples $\left\{y_{1}, \ldots, y_{K}\right\}$ from a proposal distribution $\pi(y)$ and then choosing $y$ from $\left\{y_{1}, \ldots, y_{K}\right\}$ in proportion to the importance weights $w_{y}=\frac{e^{\psi(x, y)}}{\sum_{k} e^{\psi\left(x, y_{k}\right)}}$, where $\psi$ is a function that takes $x$ and $y$ and outputs a scalar.   {According to the section 2.3 in \cite{oord2018representation}}, by setting the proposal distribution as the marginal distribution $\pi(y) \equiv p(y)$, the unnormalized density of $y$ given a specific set of samples $y_{2: K}=\left\{y_{2}, \ldots, y_{K}\right\}$ and $x$ is:}
\begin{equation}
  {
q\left(y \mid x, y_{2: K}\right)=p(y) \cdot \frac{K \cdot e^{\psi(x, y)}}{e^{\psi(x, y)}+\sum_{k=2}^{K} e^{\psi\left(x, y_{k}\right)}}}
\end{equation}
  {
where $K$ denotes the numbers of samples. 
  {According to the equation 3 of section 2 in \cite{cremer2017reinterpreting}}, the expectation of $q\left(y \mid x, y_{2: K}\right)$ with respect to resampling of the alternatives $y_{2: K}$ from $p(y)$ produces a normalized density:}
\begin{equation}
  {
    \bar{q}(y \mid x)=\mathbb{E}_{p\left(y_{2: K}\right)}\left[q\left(y \mid x, y_{2: K}\right)\right]}
\label{eq:expectation}
\end{equation}
  {
With Equation \ref{eq:expectation} and Jensen's inequality applied in Equation \ref{eq:barber-agakov-bound}, we have }
\begin{equation}
  {
\begin{aligned}
I(x, y) & \geq \mathbb{E}_{p(x, y)} \log \frac{\mathbb{E}_{p\left(y_{2: K}\right)} q\left(y \mid x, y_{2: K}\right)}{p(y)} \geq
\mathbb{E}_{p(x, y)}\left[\mathbb{E}_{p\left(y_{2: K}\right)} \log \frac{q\left(y \mid x, y_{2: K}\right)}{p(y)}\right]
\\
&=
\mathbb{E}_{p(x, y)}\left[\mathbb{E}_{p\left(y_{2: K}\right)} \log \frac{p(y) K \cdot w_{y}}{p(y)}\right] \\
&=\mathbb{E}_{p(x, y)}\left[\mathbb{E}_{p\left(y_{2: K}\right)} \log \frac{K \cdot e^{\psi(x, y)}}{e^{\psi(x, y)}+\sum_{k=2}^{K} e^{\psi\left(x, y_{k}\right)}}\right] 
\end{aligned}
}
\end{equation}

  {
It is obviously that $\frac{ e^{\psi(x, y)}}{e^{\psi(x, y)}+\sum_{k=2}^{K} e^{\psi\left(x, y_{k}\right)}} \leq 1$, thus we have}
\begin{equation}
  {
\mathbb{E}_{p(x, y)}\left[\mathbb{E}_{p\left(y_{2: K}\right)} \log \frac{K \cdot e^{\psi(x, y)}}{e^{\psi(x, y)}+\sum_{k=2}^{K} e^{\psi\left(x, y_{k}\right)}}\right] \leq \log K
}
\label{eq:3}
\end{equation}

  {
With Equation \ref{eq:3}, we have}
\begin{equation}
  {
\begin{aligned}
&\mathbb{E}_{p(x, y)}\left[\mathbb{E}_{p\left(y_{2: K}\right)} \log \frac{K \cdot e^{\psi(x, y)}}{e^{\psi(x, y)}+\sum_{k=2}^{K} e^{\psi\left(x, y_{k}\right)}}\right] \\
=& \mathbb{E}_{p\left(x, y_{1}\right) p\left(y_{2: K}\right)}\left[\log \frac{e^{\psi(x, y)}}{\frac{1}{K} \sum_{k=1}^{K} e^{\psi\left(x, y_{k}\right)}}\right]=I_{\text{NCE}}(x ; y \mid \psi, K) \leq \log K,
\end{aligned}}
\end{equation}

  {Therefore, we have }
\begin{equation}
  {
I(x, y) \geq I_{\text{NCE}}(x ; y \mid \psi, K) \leq \log K}
\end{equation}

  {
If $I(x,y)>\log K$, then $I(x, y)>\log K \geq I_{\text{NCE}}(x ; y \mid \psi, K)$, and $I_{\text{NCE}}$ will be a loose bound.}

  {
Thus, $I_{\text{NCE}} \leq I(x,y) \leq \log K$ is the necessary condition for $I_{\text{NCE}}$ to be a tight bound of $I(x,y)$.}

\subsection{  {Detailed derivation of Theorem 1}}
  {
As the number of confounders increases, although the true mutual information $I(c ; \mathcal{T})$ does not increase, the necessary condition of $I_{NCE}$ to be a tight lower bound of $I_{NCE}$ becomes more difficult to satisfy, and the demand of data  increases significantly.}


  {As for an entangled context, the
necessary condition of the InfoNCE lower bound $I_{NCE}(c; \mathcal{T})$ to be a tight bound is }

\begin{equation}
  {
  I_{N C E}\left(c ; \mathcal{T}\right) \leq I\left(c ; \mathcal{T}\right) \leq \log K  }
\end{equation}

  {
Since $I(c ; \mathcal{T}) \geq \sum_{i=0}^{N} I\left(u_{i} ; \mathcal{T}\right)$, to let the above condition satisfied, the amount of data $K$ must satisfy}

\begin{equation}
      {\log K \geq \sum_{i=0}^{N} I\left(u_{i} ; \mathcal{T}\right)}
\end{equation}

\begin{equation}
      {K \geq e^{\sum_{i=0}^{N} I(u_{i} ; \mathcal{T})}}
\end{equation}

  {
Therefore, if the number of confounders increases, then the demand for data will grow exponentially.}

  {
When data is not rich enough, the nesseray condition may not be satisfied. The InfoNCE lower bound $I_{NCE}(c ; \mathcal{T})$ may be loose, that is $I_{NCE}(c ; \mathcal{T})$ may be much smaller than the true mutual information $I(c ; \mathcal{T})$, thus the MI optimization based on $I_{NCE}(c ; \mathcal{T})$ will be severely affected.}

  {
$I_{N C E}\left(c_{i} ; \mathcal{T}\right)$ is the lower bound of $I\left(c_{i} ; \mathcal{T}\right)$ and the necessary condition of $I_{N C E}\left(c_{i} ; \mathcal{T}\right)$ to be a tight bound of  $I\left(c_{i} ; \mathcal{T}\right)$ is}

\begin{equation}
      {I_{N C E}\left(c_{i} ; \mathcal{T}\right) \leq I\left(c_{i} ; \mathcal{T}\right) \leq \log K}
\end{equation}

  {
As for disentangled context  $c=\{c_{1},c_{2},\cdots,c_{N}\}$, we then derive the necessary condition of $I(c,\mathcal{T})$ to be a tight lower bound of $I(c,\mathcal{T})$:}

  {
With the assumption that the contexts $\{c_{1},c_{2},\cdots,c_{N}\}$ are independent to each other, then $I(c ; \mathcal{T})$ could be derived as $\sum I\left(c_{i} ; \mathcal{T}\right)$. Therefore, under the confounder independent assumption, let $I_{NCE}(c ; \mathcal{T})$ be a tight bound is only necessary to let every $I_{NCE}(c_{i} ; \mathcal{T})$ to be a tight bound. }

  {
If every $I_{NCE}(c_{i} ; \mathcal{T})(i=1,2,\ldots,N)$ is a tight bound, then we have}

\begin{equation}
      {I_{N C E}\left(c_{i} ; \mathcal{T}\right) \leq I\left(c_{i} ; \mathcal{T}\right) \leq \log K}
\end{equation}

  {
under the confounder independent assumption, we have}

\begin{equation}
      {\sum I_{N C E}\left(c_{i} ; \mathcal{T}\right) \leq   \sum I\left(c_{i} ; \mathcal{T}\right) \leq N \log K}
\end{equation}

\begin{equation}
      {I_{N C E} \left(c ; \mathcal{T}\right) = \sum I_{N C E}\left(c_{i} ; \mathcal{T}\right) \leq I\left(c ; \mathcal{T}\right) = \sum I\left(c_{i} ; \mathcal{T}\right) \leq N \log K}
\end{equation}

  {
Thus, the necessary condition of  $I_{N C E} \left(c ; \mathcal{T}\right)$ to be a tight bound of $I \left(c ; \mathcal{T}\right)$ could be relaxed to} 

\begin{equation}
      {I_{N C E} \left(c ; \mathcal{T}\right)  \leq I\left(c ; \mathcal{T}\right)  \leq N \log K}
\end{equation}

  {
Therefore, by decomposing the MI estimation under the confounder independent assumption, the demand of the amount $K$ of data could be reduced from  $K \geq e^{I(c ; \mathcal{T})}$ to $K \geq e^{\frac{1}{N} I(c; \mathcal{T})}$. And with $I(c ; \mathcal{T}) \geq \sum_{i=0}^{N} I\left(u_{i} ; \mathcal{T}\right)$, specificly, the the amount $K$ of data could be reduced from  $K \geq e^{\sum_{i=0}^{N} I\left(u_{i} ; \mathcal{T}\right)}$ to $K \geq e^{\frac{1}{N} \sum_{i=0}^{N} I\left(u_{i} ; \mathcal{T}\right)}$.}

\section{Pseudo-code}


\subsection{Combination with model-based methods}
\label{model-based pseudo-code}
We provide the pseudo-code of \modelname combined with model-based methods. Firstly, the past state-action pairs are encoded into the disentangled context vectors by the context encoder. According to the learned context, the transition prediction network predicts the future states of different actions. 
Then, the context encoder is optimized by maximizing the mutual information between the disentangled context vectors and historical trajectories while minimizing the state transition prediction error.
In particular, we use the cross entropy method (CEM) \cite{botev2013cross}, a typical neural model predictive control (MPC) \cite{garcia1989model} method, to select actions, in which several candidate action sequences are iteratively sampled from a candidate distribution, which is adjusted based on best-performing action samples. 
\begin{algorithm}[htbp]

\caption{Training \modelname with context-aware world model}
\label{alg:training}
\begin{algorithmic}[0]
\STATE {\bf Inputs}: learning rate $\alpha$,maximum number of iteration $P$, batch size $B$, the number of past observations $H_{\text{past}}$, maxium rollout step $max\_step$ and  the number of future observations $H_{\text{future}}$.
\STATE Initialize parameters of prediction network $\phi$,  context encoder $\varphi$. 
\STATE Initialize replay buffer $\mathcal{D} \leftarrow \emptyset$.
\FOR {$P$ iterations}
\STATE {{\textsc{// Collect training samples}}}
\STATE $step=0$
\STATE $\mathcal{V}=0$
\WHILE{$step$ $\leq$ $max\_step$}
\STATE Sample $u_{\mathcal{V}}\sim p_{u_{\tt train}}\left(u\right)$. 
\STATE $\mathcal{V} =\mathcal{V} + 1$
\FOR{$t=1$ {\bfseries to} TaskHorizon}  
\STATE $step = step +1$
\STATE Get context latent vectors $c_{t_{0}}, c_{t_{1}},\ldots,c_{t_{N}}= g\left(\tau_{t}; \varphi\right), \tau_{t}=\left\{s_{l},a_{l}\right\}_{l=t-H_{\text{past}}}^{t-1}$
\STATE Collect samples $\{(s_t,a_t,s_{t+1},r_t,\tau_{t})\}$ from the environment 
using the planning algorithm based on CEM with the context vectors
\ENDFOR
\STATE Update $\mathcal{D}_{u_{\mathcal{V}}} \leftarrow \mathcal{D}_{u_{\mathcal{V}}} \cup \{(s_t,a_t,s_{t+1},r_t,\tau_{t})\}$
\ENDWHILE
\STATE {{\textsc{// Update dynamics models and encoder}}}
\STATE Initialize batch $\mathcal{B} \leftarrow \emptyset$.
\FOR{$i=1$ {\bfseries to} $B$}
\STATE sample $\mathcal{V}^{*}$ from [0,$\mathcal{V}_{max}$]
\STATE Sample $\left\{s_t,a_t,s_{t+1},r_t,\tau_{t}\right\}$ from $\mathcal{D}_{u_{\mathcal{V}^{*}}}$
\STATE Sample positive trajectories $\tau^{+}$ from $\mathcal{D}_{u_{\mathcal{V}^{*}}}$
\STATE Sample negative trajectories $\left\{\tau_{k}^{-}\right\}_{k=2}^{K}$ from $\mathcal{D}_{u_{\mathcal{V} !=\mathcal{V}^{*}}}$
\STATE Get context latent vectors $c_{t_{0}}, c_{t_{1}},\ldots,c_{t_{N}}= g\left(\tau_{t}; \varphi\right)$
\STATE Update $\mathcal{B} \leftarrow \mathcal{B}\cup \{(s_t,a_t,s_{t+1},r_t,\tau_{t})\}$
\ENDFOR
\STATE $\scriptsize{\mathcal{L}^{\tt pred} \leftarrow 
E_{\tau^{*} \sim \mathcal{B}}\left[-\frac{1}{H} \sum_{\lambda=t}^{t+H_{\text{future}}-1} \log f_{\phi}\left(s_{\lambda+1} \mid s_{\lambda}, a_{\lambda},\left(c_{0_{\lambda}}, \ldots, c_{N_{\lambda}}\right)\right)\right]}$ 
\STATE $\mathcal{L}^{\mathrm{NCE}} \leftarrow  \sum_{i}^{N} I_{\mathrm{NCE}}\left(c_{i} ; \mathcal{T}\right)-\sum_{j}^{N} \sum_{i=0, i \neq j}^{N} I_{\mathrm{NCE}}\left(c_{i} ; c_{j}\right)$

\STATE Update
$\varphi \leftarrow \varphi - \alpha \nabla_{\varphi}
\mathcal{L}^{\tt NCE}$

\STATE Update
$\varphi \leftarrow \varphi - \alpha \nabla_{\varphi}
\mathcal{L}^{\tt pred}$

\STATE Update
$\phi \leftarrow \phi - \alpha \nabla_{\phi}
\mathcal{L}^{\tt pred}$

\ENDFOR
\end{algorithmic}
\end{algorithm}

\newpage
\subsection{Combination with model-free methods}
\label{model-free pseudo-code}
We provide the pseudo-code for the combination between \modelname and the model-free method, which uses the context encoder learned by \modelname as a plug-and-play module to extract accurate context. 
We concatenate the disentangled context encoded by a pre-trained context encoder from \modelname and the current state-action pairs, and learn a conditional policy $\pi \left(a_t|s_t,c_{0},\ldots,c_{N}\right)$.
We choose the Proximal Policy Optimization (PPO) method \cite{schulman2017proximal} to train the agents. 
\begin{algorithm}
\caption{Proximal Policy Optimization with disentangled context encoder learned by DOMINO}
\label{alg:PPO}
\begin{algorithmic}
\STATE {\bf Inputs}:Maximum number of iteration $P$, number of actor updates $M$, number of critic updates $B$, the KL regular coefficient $\lambda$, scaling coefficient $\alpha$ and the learning rate $\beta$.
\STATE Initialize parameters of policy network $\theta$, value network $\xi$, and context encoder $\varphi$.
    \FOR {$P$ iterations}
        \STATE Encode disentangled context vectors $c_{t_{0}},c_{t_{1}},\ldots,c_{t_{N}}$ by the learned context encoder $g_{\varphi}(\cdot)$
        \STATE Run policy $\pi_{\theta}$ for $T$ timesteps, collecting 
        $\{\{s_t,c_{t_{0}},c_{t_{1}},\ldots,c_{t_{N}},a_t,r_t\}\}_{t=1}^{T}$
        \STATE Estimate advantages $\hat{A}_t = \sum_{t' > t} \gamma^{t'-t} r_{t'} - V_\xi(s_t,c_{t_{0}},c_{t_{1}},\ldots,c_{t_{N}})$
        \STATE $\pi_\mathrm{old} \leftarrow \pi_\theta$
        \FOR {$M$ updates}
            \STATE $J_{\text{PPO}}(\theta) \leftarrow
            -\left\{
            \sum_{t=1}^T \frac{\pi_{\theta}(a_t|s_t,c_{t_{0}},c_{t_{1}},\ldots,c_{t_{N}})}{\pi_{old}(a_t|s_t,c_{t_{0}},c_{t_{1}},\ldots,c_{t_{N}})} \hat{A}_t - \lambda \mathrm{KL}[\pi_{old}|\pi_{\theta}]\right\} $
            \STATE $\theta \leftarrow \theta -\beta \nabla_{\theta} J_{\text{PPO}}$
		\ENDFOR
		\FOR {$B$ updates}
            
            \STATE $L_{\text{BL}}(\xi) \leftarrow  \sum_{t=1}^T (\sum_{t' > t} \gamma^{t'-t} r_{t'} - V_\xi(s_t,c_{t_{0}},c_{t_{1}},\ldots,c_{t_{N}}))^2$ 
            \STATE $\xi \leftarrow \xi - \beta \nabla_{\xi}L_{\text{BL}}$
		\ENDFOR
		\IF {$\mathrm{KL}[\pi_{old}|\pi_{\theta}] > \beta_{\text{high}}\mathrm{KL}_{\text{target}}$}
		\STATE  $\lambda \leftarrow \alpha \lambda$
		\ELSIF {$\mathrm{KL}[\pi_{old}|\pi_{\theta}] < \beta_{\text{low}}\mathrm{KL}_{\text{target}}$}
		\STATE  $\lambda \leftarrow  \lambda/\alpha $
		\ENDIF
	\ENDFOR	
\end{algorithmic}
\end{algorithm}

\section{Details about the testing environments}

\label{Detail env}
\setcounter{table}{2}
\begin{table}[t]
\centering
\caption{Environment parameters used for the multi-confounded  experiments.}
\begin{adjustbox}{max width=0.85\textwidth}
\renewcommand\arraystretch{1.5}
\begin{tabular}{@{}crlrlrlc@{}}
\toprule[0.5mm]
& \multicolumn{2}{c}{Train}                      
& \multicolumn{2}{c}{\begin{tabular}[c]{@{}c@{}}Test\end{tabular}}\\ \toprule[0.5mm]
\multicolumn{1}{c}{\multirow{2}{*}{CartPole}}     & $f\in$  & \multicolumn{1}{l}{\begin{tabular}[c]{@{}l@{}}$\{5.0, 6.0, 7.0, 8.0, 9.0, 10.0,$\\ $\hspace{1mm} 11.0, 12.0, 13.0, 14.0, 15.0\}$\end{tabular}}    & $f\in$       & \multicolumn{1}{l}{$\{3.0, 3.5, 16.5, 17.0\}$} \\
\multicolumn{1}{c}{}                              & $l\in$ & \multicolumn{1}{l}{$\{0.40, 0.45, 0.50, 0.55, 0.60\}$}                                                                                            & $l\in$       & \multicolumn{1}{l}{$\{0.25, 0.30, 0.70, 0.75\}$}\\ 
\bottomrule[0.1mm]
\multicolumn{1}{c}{\multirow{2}{*}{Pendulum}}     & $m\in$  & \multicolumn{1}{l}{\begin{tabular}[c]{@{}l@{}}$\{0.75, 0.80, 0.85, 0.90, 0.95,$\\ $\hspace{1mm}1.0, 1.05, 1.10, 1.15, 1.20, 1.25\}$\end{tabular}} & $m\in$       & \multicolumn{1}{l}{$\{0.50, 0.70, 1.30, 1.50\}$}\\
\multicolumn{1}{c}{}                              & $l\in$  & \multicolumn{1}{l}{\begin{tabular}[c]{@{}l@{}}$\{0.75, 0.80, 0.85, 0.90, 0.95,$\\ $\hspace{1mm}1.0, 1.05, 1.10, 1.15, 1.20, 1.25\}$\end{tabular}} & $l\in$       & \multicolumn{1}{l}{$\{0.50, 0.70, 1.30, 1.50\}$}\\
\bottomrule[0.1mm]
\multicolumn{1}{c}{\multirow{2}{*}{Half-cheetah}} & $m\in$  & \multicolumn{1}{l}{$\{0.75, 0.85, 1.0, 1.15, 1.25\}$}                                                                                             & $m\in$       & \multicolumn{1}{l}{$\{0.40, 0.50, 1.50, 1.60\}$}\\
\multicolumn{1}{c}{}                              & $d\in$  & \multicolumn{1}{l}{$\{0.75, 0.85, 1.0, 1.15, 1.25\}$}                                                                                             & $d\in$       & \multicolumn{1}{l}{$\{0.40, 0.50, 1.50, 1.60\}$}\\
\bottomrule[0.1mm]
\multicolumn{1}{c}{\multirow{2}{*}{Ant}} & $m\in$  & \multicolumn{1}{l}{$\{0.75,0.85,1.0,1.15,1.25\}$}                                                                                             & $m\in$       & \multicolumn{1}{l}{$\{0.40, 0.50, 1.50, 1.60\}$}\\
\multicolumn{1}{c}{}                              & $d\in$  & \multicolumn{1}{l}{$\{0.75,0.85,1.0,1.15,1.25\}$}                                                                                             & $d\in$       & \multicolumn{1}{l}{$\{0.40, 0.50, 1.50, 1.60\}$}\\ 
\bottomrule[0.1mm]
\multicolumn{1}{c}{\multirow{2}{*}{SlimHumanoid}} & $m\in$  & \multicolumn{1}{l}{$\{0.80, 0.90, 1.0, 1.15, 1.25\}$}                                                                                             & $m\in$       & \multicolumn{1}{l}{$\{0.60, 0.70, 1.50, 1.60\}$}\\
\multicolumn{1}{c}{}                              & $d\in$  & \multicolumn{1}{l}{$\{0.80, 0.90, 1.0, 1.15, 1.25\}$}                                                                                             & $d\in$       & \multicolumn{1}{l}{$\{0.60, 0.70, 1.50, 1.60\}$}\\ 
\bottomrule[0.1mm]
\multicolumn{1}{c}{\multirow{2}{*}{Crippled Ant}} & $m\in$  & \multicolumn{1}{l}{$\{0.75,0.85,1.0,1.15,1.25\}$}                                                                                             & $m\in$       & \multicolumn{1}{l}{$\{0.40, 0.50, 1.50, 1.60\}$}\\
\multicolumn{1}{c}{}                              & $d\in$  & \multicolumn{1}{l}{$\{0.75,0.85,1.0,1.15,1.25\}$}                                                                                             & $d\in$       & \multicolumn{1}{l}{$\{0.40, 0.50, 1.50, 1.60\}$}\\ 
\multicolumn{1}{c}{}                              &   \multicolumn{2}{l}{crippled leg:$\{0,1,2\}$}            &         \multicolumn{2}{l}{crippled leg:$\{3\}$}\\ 
\bottomrule[0.1mm]
\multicolumn{1}{c}{\multirow{2}{*}{Crippled Halfcheetah}} & $m\in$  & \multicolumn{1}{l}{$\{0.75,0.85,1.0,1.15,1.25\}$}                                                                                             & $m\in$       & \multicolumn{1}{l}{$\{0.40, 0.50, 1.50, 1.60\}$}\\
\multicolumn{1}{c}{}   & $d\in$  & \multicolumn{1}{l}{$\{0.75,0.85,1.0,1.15,1.25\}$}                                                                                             & $d\in$       & \multicolumn{1}{l}{$\{0.40, 0.50, 1.50, 1.60\}$}\\
\multicolumn{1}{c}{}                              &   \multicolumn{2}{l}{crippled leg:$\{0\}$}            &         \multicolumn{2}{l}{crippled leg:$\{1\}$}\\ 
\toprule[0.5mm]
\end{tabular}
\end{adjustbox}
\label{tbl:environment}
\end{table}
\definecolor{codegreen}{rgb}{0,0.8,0}
\definecolor{codegray}{rgb}{0.5,0.5,0.5}
\definecolor{codepurple}{rgb}{0.58,0,0.82}
\definecolor{backcolour}{rgb}{0.99,0.99,0.97}
\begin{lstlisting}[
float=htbp,
language=Python,
floatplacement=htbp,
xleftmargin=2em,
frame=single,
framexleftmargin=1.5em,
backgroundcolor=\color{backcolour},
belowskip=-1\baselineskip,
commentstyle=\color{codegreen},
keywordstyle=\color{magenta},
numberstyle=\tiny\color{codegray},
stringstyle=\color{codepurple},
backgroundcolor=\color{backcolour},
commentstyle=\color{codegreen},
basicstyle=\ttfamily\scriptsize,
breakatwhitespace=false,         
numbers=left,                    
breaklines=true,                 
captionpos=b,                    
keepspaces=true,                 
numbersep=5pt,                  
showspaces=false,                
showstringspaces=false,
showtabs=false,                  
tabsize=2,
label={lst:code},
caption=PyTorch-style pseudo-code for dynamics change based on Mujoco engine.]
def change_env(self):
        mass = np.copy(self.original_mass)
        damping = np.copy(self.original_damping)
        mass *= self.mass_scale
        damping *= self.damping_scale
        self.model.body_mass[:] = mass
        self.model.dof_damping[:] = damping
\end{lstlisting}
\begin{lstlisting}[
float=t,
language=Python,
floatplacement=t,
xleftmargin=2em,
frame=single,
framexleftmargin=1.5em,
backgroundcolor=\color{backcolour},
belowskip=-1\baselineskip,
commentstyle=\color{codegreen},
keywordstyle=\color{magenta},
numberstyle=\tiny\color{codegray},
stringstyle=\color{codepurple},
backgroundcolor=\color{backcolour},
commentstyle=\color{codegreen},
basicstyle=\ttfamily\scriptsize,
breakatwhitespace=false,         
numbers=left,                    
breaklines=true,                 
captionpos=b,                    
keepspaces=true,                 
numbersep=5pt,                  
showspaces=false,                
showstringspaces=false,
showtabs=false,                  
tabsize=2,
label={lst:code},
caption=PyTorch-style pseudo-code for multi-confounded environments initialization.]
def reset_model(self):
        c = 0.01
        self.set_state(
            self.init_qpos + self.np_random.uniform(low=-c, high=c, size=self.model.nq),
            self.init_qvel + self.np_random.uniform(low=-c, high=c, size=self.model.nv,)
        )
        pos_before = mass_center(self.model, self.sim)
        self.prev_pos = np.copy(pos_before)

        random_index = self.np_random.randint(len(self.mass_scale_set))
        self.mass_scale = self.mass_scale_set[random_index]

        random_index = self.np_random.randint(len(self.damping_scale_set))
        self.damping_scale = self.damping_scale_set[random_index]

        self.change_env()
        return self._get_obs()
        \end{lstlisting}
For CartPole environments, we use open-source implementation of
CartPoleSwingUp-v2\footnote{\scriptsize{We use implementation available at https://github.com/0xangelo/gym-cartpole-swingup}}, which is the modified version of original CartPole environments from OpenAI Gym. The objective of CartPole task is to swing up the pole by moving a cart and keep the pole upright. For our experiments, we modify  the push force $f$ and the pole length $l$ simultaneously.
As for Pendulum, we scale the pendulum mass by scale factor $m$ and modify the  pendulum length $l$.
For Pendulum environments, we use the open-source implementation of from the OpenAI Gym. The objective of Pendulum is to swing up the pole and keep the pole upright within 200 timesteps. We scale the pendulum mass by scale factor $m$ and modify the  pendulum length $l$.

As for Hopper, Half-cheetah, Ant, and Slimhumanoid, we use the  environments from MuJoCo physics engine \footnote{\scriptsize{We use implementation available at https://github.com/iclavera/learning\_to\_adapt}}, and scale the mass of every rigid link by scale factor $m$, and scale damping of every joint by  scale factor $d$.
As for Crippled Ant and Crippled Half-cheetah, we scale the mass of every rigid link by scale factor $m$, scale damping of every joint by  scale factor $d$, and randomly select one leg, and make it crippled. The objectives of these tasks are to move forward as fast as possible while minimizing the action cost.
The detailed settings are illustrated in Table \ref{tbl:environment}.  We provide the pyTorch-style pseudo-code for multi-confounded environments in Listing 1 and Listing 2.
We implement these environments based on the publicly available code provide by \cite{nagabandi2018learning,seo2020trajectory}, and we also open-source the code  of the \href{https://anonymous.4open.science/r/Multiple-confounded-Mujoco-Envs-01F3}{multiple-confounded environments}$\footnote{\scriptsize{We provide open-source environments at 
https://anonymous.4open.science/r/Multiple-confounded-Mujoco-Envs-01F3}}$. 
For both the training and testing phase, we sample the confounders at the beginning of each episode. During training, we randomly 
select a combination of confounders from a training set. At test time, we evaluate each algorithm in unseen environments with confounders outside the training range. 
We also provide the PyTorch-style pseudo-code for the dynamics change based on Mujoco engine.

\section{Implementation details}
\label{app_detail}
\subsection{Combination with Model-based RL}
The context encoder is modeled as multi-layer perceptrons (MLPs) with 3 hidden layers and N output heads which are single-layer MLPs. Every disentangled context vector is produced as a 10-dimensional vector by the 3 hidden layers and a specific output head.
 Then, the disentangled context vectors are used as the additional input to the prediction network, i.e., the input is given as a concatenation of state, action, and context vector. 
We use $H_{past}=10$ for the number of past observations and $H_{future} =5$ for the number of future observations. The prediction network is modeled as multi-layer perceptrons (MLPs) with 4 hidden layers of 200 units each and Swish activations. For each prediction head, the mean and variance are parameterized by a single linear layer that takes the output vector of the backbone network as an input.
To train the prediction network, we collect 10 trajectories with 200 timesteps from environments using the MPC controller and train the model for 50 epochs at every iteration. We train the prediction network for 10 iterations for every experiment. We evaluate trained models on environments over 8 random seeds every iteration to report the testing performance. The Adam optimizer \cite{kingma2014adam} is used with a learning rate $1\times 10^{-4}$. For planning, we use the cross entropy method (CEM) with 200 candidate actions for all the environments. The horizon of MPC is set as 30.

\subsection{Combination with Model-free RL}
We train the model-free agents for 5 million timesteps on OpenAI-Gym and MuJoCo environments (i.e., Hopper, Half-cheetah, Ant, Crippled Half-cheetah, Crippled Half-Ant, Slim-Humanoid) and 0.5 million timesteps on CartPole and Pendulum. 
The trained agents are evaluated every 10,000 timesteps over 5 random seeds. We use a discount factor $\gamma = 0.99$, a generalized advantage estimator \cite{schulman2015high} parameter $\lambda = 0.95$ and an entropy bonus of 0.01 for exploration. 
In every iteration, the agent rollouts 200 timesteps in the environments with the learned policy,  and then it will be trained for 8 epochs with 4 mini-batches. The Adam optimizer is used with the learning rate $5 \times 10^{-4}$.

\subsection{Details of InfoNCE}
We provide detailed pseudocode for the calculation of InfoNCE bound. Specifically, the temperature $\tau$ is set as 0.004 to the calculation of $I(c_{i},\mathcal{T})$ and is set as 0.1 to calculation of $I(c_{i},c_{j})$.

\begin{algorithm}[htbp]
\caption{Pseudocode of InfoNCE  in a PyTorch-like style.}
\label{alg:code}
\definecolor{codeblue}{rgb}{0.25,0.5,0.5}
\lstset{
  backgroundcolor=\color{white},
  basicstyle=\fontsize{7.2pt}{7.2pt}\ttfamily\selectfont,
  columns=fullflexible,
  breaklines=true,
  captionpos=b,
  commentstyle=\fontsize{7.2pt}{7.2pt}\color{codeblue},
  keywordstyle=\fontsize{7.2pt}{7.2pt},
}
\begin{lstlisting}[language=python]
# x_q input vector
# x_k positive sample
# x_que negative samples
# f_q, f_k, f_que: encoder networks for query, key and queue
# m: momentum
# t: temperature

def InfoNCE(x_q,x_k,x_que):
    q = f_q.forward(x_q)  # queries: NxC
    k = f_k.forward(x_k)  # keys: NxC
    k = k.detach()  # no gradient to keys
    queue = f_que.forward(x_que) #  keys: Cx(K-1)

    # positive logits: Nx1
    l_pos = bmm(q.view(N,1,C), k.view(N,C,1))

    # negative logits: Nx(K-1)
    l_neg = mm(q.view(N,C), queue.view(C,K-1))

    # logits: NxK
    logits = cat([l_pos, l_neg], dim=1)

    # contrastive loss, Eqn.(1)
    labels = zeros(N)  # positives are the 0-th
    loss = CrossEntropyLoss(logits/t, labels)
\end{lstlisting}
\end{algorithm}

\subsection{  {Details of the adaptive planning used in adaption process}}
  {
\label{adaptive planning}
The prediction model has 3 output head $head_{0},head_{1},head_{2}$ which are used for selecting actions by planning. The adaptive planning method selects the most accurate prediction head over a recent experience. Given $N$ past transitions, we select the prediction head $h_{*}$ by
$$\underset{head \in[Head]}{\operatorname{argmin}} \sum_{i=t-N}^{t-2} \ell\left(s_{i+1}, f\left(s_{i+1} \mid s_{i}, a_{i},\left(c_{{0}_{\lambda}}, \ldots, c_{{N}_{\lambda}}\right);\phi,head\right)\right)$$
where $\ell$ is the mean square error function. All the hyper-parameter is set as same as T-MCL\cite{lee2020context}.}

\section{Additional Results}
\label{add_results}
\subsection{Prediction Error}
\begin{figure}[htbp]
    \centering
    \includegraphics[width=0.99\linewidth]{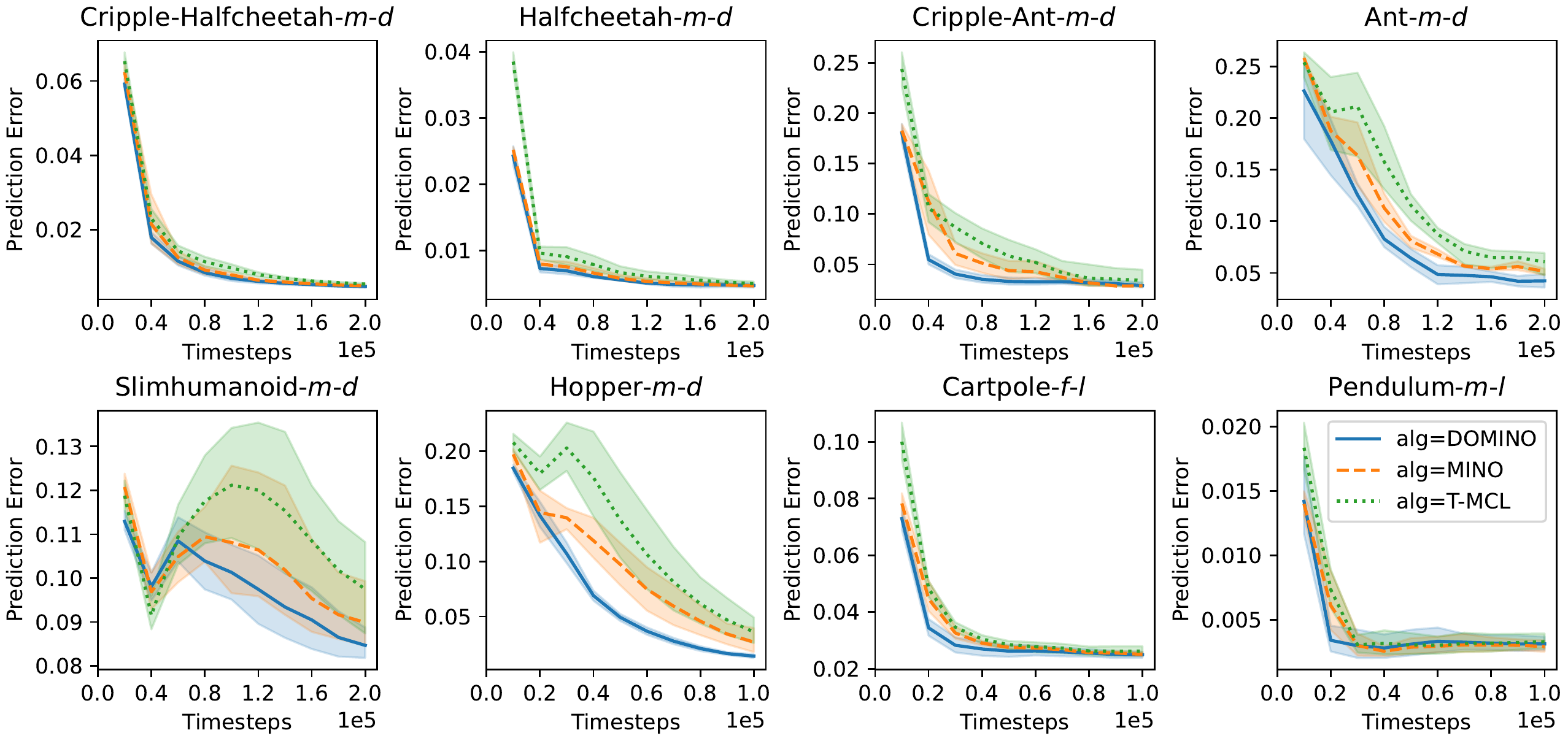}
    \caption{
Comparison with the \textit{model-based methods} of the Prediction Error. The results show the mean and standard deviation of average returns averaged over 8 runs.}
\label{fig:mb_pre_error_app}
\end{figure}

As shown in Figure \ref{fig:mb_pre_error_app}, \modelname has a smaller prediction error compared to T-MCL and its ablation version MINO (optimize MI with entangled context), indicating that the learned context can effectively help predict the future state more accurately, which is the key to the performance of the model-based planning.
\label{app-add}
\begin{figure}[htbp]
    \centering
    \includegraphics[width=0.65 \linewidth]{ 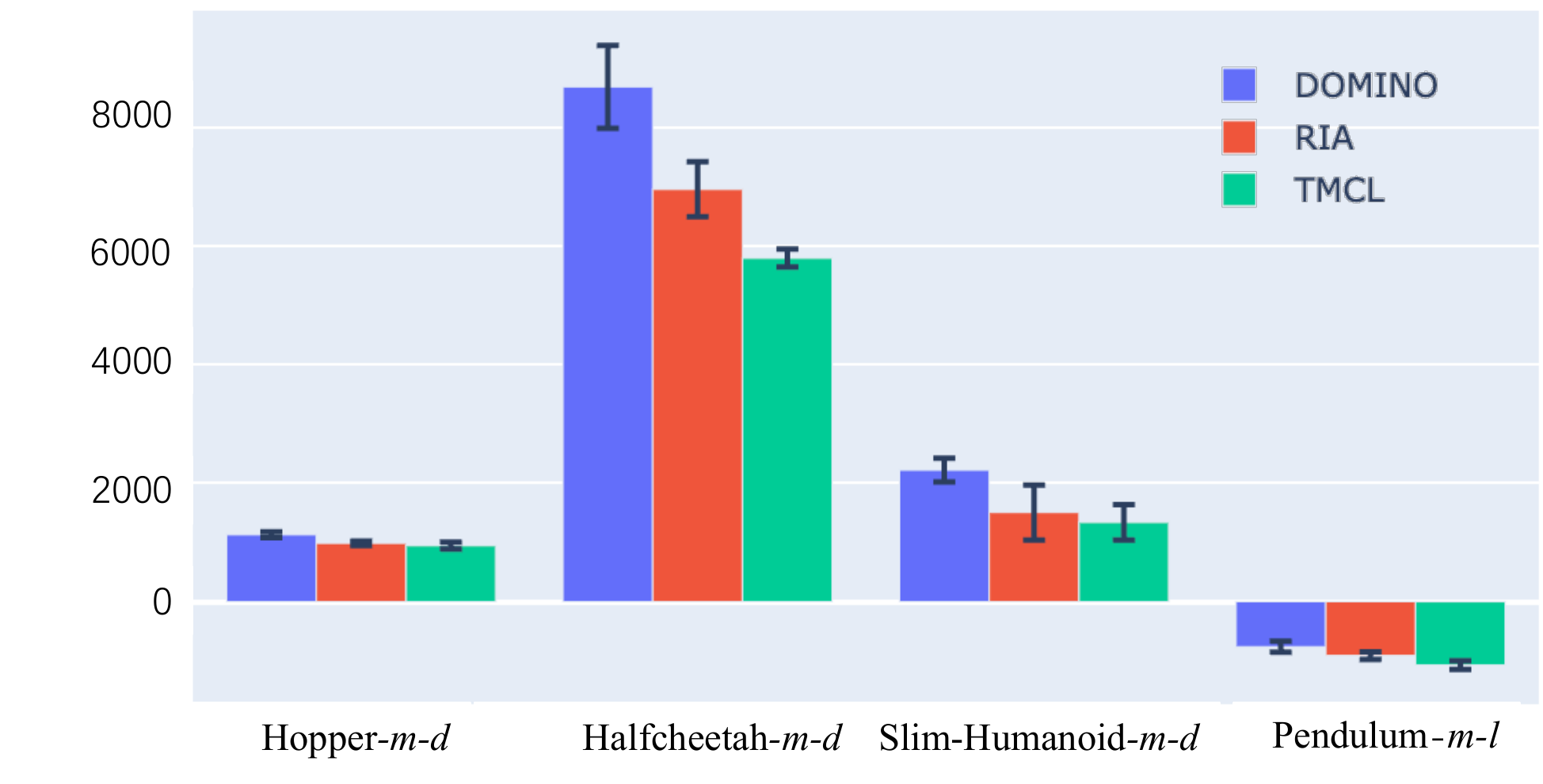}
    \caption{
  {Generalization performance comparison between DOMINO, RIA and TMCL over 5 runs (DOMINO and T-MCL are with adaptive planning).}}
\label{fig:ria-comp}
\end{figure}
\subsection{More results on the comparison with RIA}
\label{ria-compare}
  {We provide the comparison between the DOMINO without adaptive planning and RIA in the main paper. Here, we compare DOMINO and T-MCL with adaptive planning  with RIA under multi-confounded setting, the environments including Hopper-$m$-$d$, Halfcheetah-$m$-$d$, Slim-humanoid-$m$-$d$ and Pendulum-$m$-$l$. As shown in Figure \ref{fig:ria-comp}, DOMINO also achieves better generalization performance than RIA and the TMCL with adaptive planning.}

\subsection{  {Sensitivity Analysis of the hyper-parameter N}}
\begin{figure}[hbtp]
     \centering
     
     \begin{subfigure}{0.42\textwidth}
         \centering
         \includegraphics[width=\linewidth]{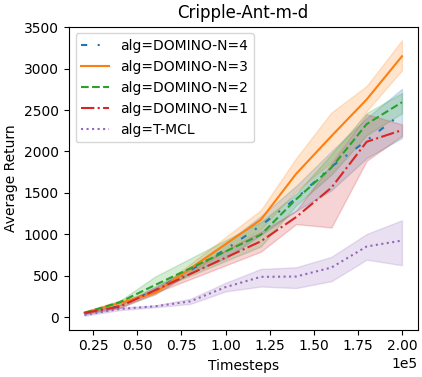}
           \label{fig:boat1}
  \caption{Performance evaluation in seen environments}
     \end{subfigure}
     \hfill
     \begin{subfigure}{0.42\textwidth}
         \centering
         \includegraphics[width=\linewidth]{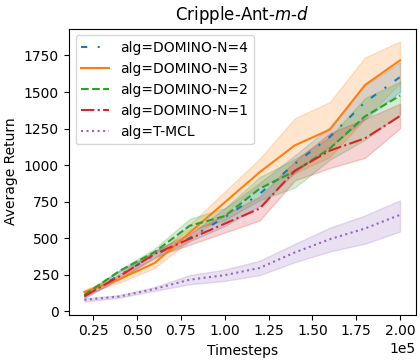}
  \caption{Performance evaluation in unseen environments}
  \label{fig:boat1}
     \end{subfigure}
        \label{fig:three graphs}
        \caption{  {The ablation of different N in Crippled-Ant-m-d domain(contains 3 confounders).}}
        \label{sense}
\end{figure}

  {
We compare the performance of DOMINO with different hyper-parameter $N$, which is equal or not equal to the number of confounders in the environment. In this experiment, the confounder is the damping, mass, and a crippled leg (number of confounders is 3), and we compare the performance of DOMINO with different hyper-parameter $N={1,2,3,4}$. 
As shown in Figure \ref{sense}, even though the hyper-parameter $N$ is not equal to the ground truth value of the confounder number, DOMINO also benefits the context learning compared to the baselines like TMCL.}





\section{Visualization}
\begin{figure}[t]
     \centering
     \begin{subfigure}{0.45\textwidth}
         \centering
         \includegraphics[width=\textwidth]{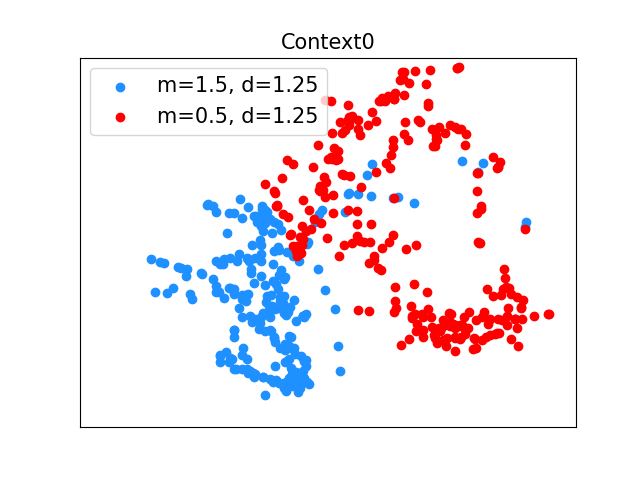}
         \caption{\footnotesize Visualization of Context0}
     \end{subfigure}
     \hfill
     \begin{subfigure}{0.45\textwidth}
         \centering
         \includegraphics[width=\textwidth]{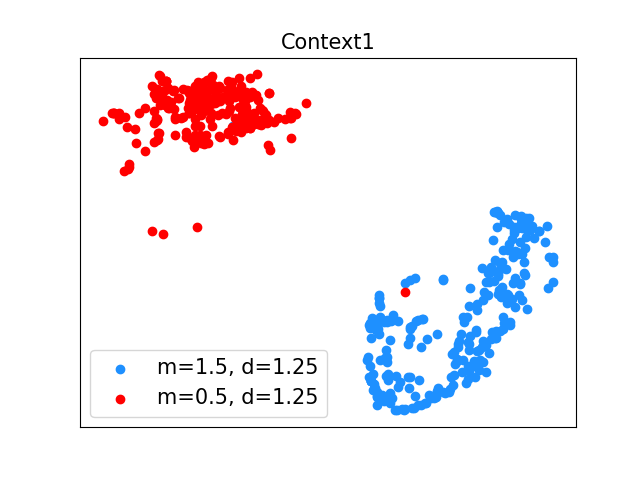}
         \caption{\footnotesize Visualization of Context1 }
     \end{subfigure}
        \caption{  {Visualization of disentangled context in with same damping scale $d=1.25$ and different mass scale $m=1.5,m=0.5$.}}
\label{fig:same damping}
\end{figure}

\begin{figure}[t]
     \centering
     \begin{subfigure}{0.45\textwidth}
         \centering
         \includegraphics[width=\textwidth]{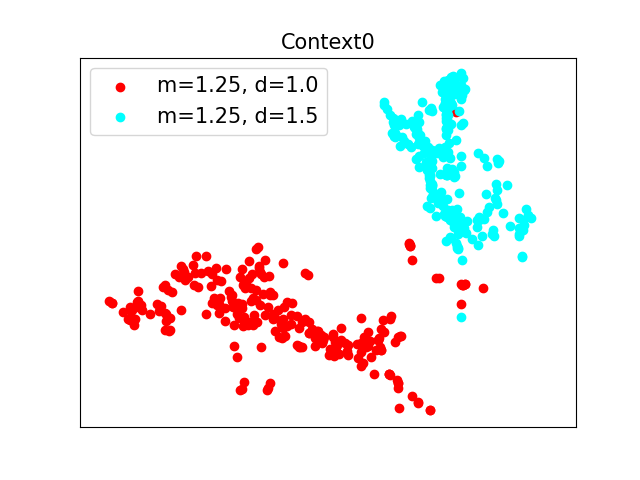}
         \caption{\footnotesize Visualization of Context0}
         \label{fig:y equals x}
     \end{subfigure}
     \hfill
     \begin{subfigure}{0.45\textwidth}
         \centering
         \includegraphics[width=\textwidth]{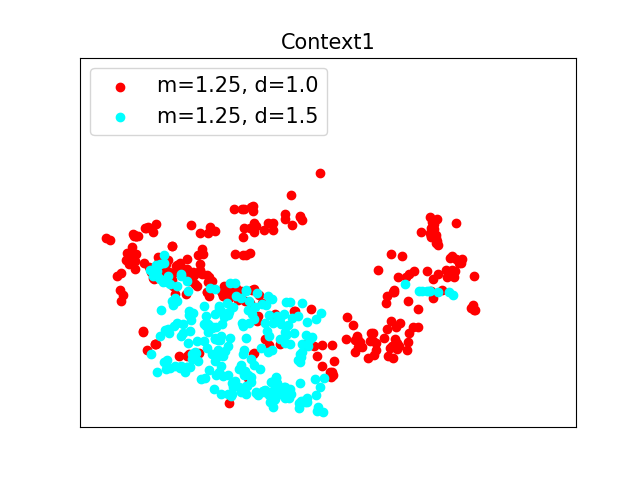}
         \caption{\footnotesize Visualization of Context1 }
         \label{fig:three sin x}
     \end{subfigure}
        \caption{  {Visualization of disentangled context in with same mass scale $m=1.25$ and different damping scale $d=1.0, d=1.5$.}}
        \label{fig: same mass}
\end{figure}

  {
\subsection{Verifying whether the contexts is disentangled}
We add an additional experiment to show that the context vectors inferred by DOMINO are disentangled well. We vary only one of the confounders and observe the changes of $N$ disentangled vectors.  In this experiment, we set up two different confounders:  mass $m$ and damping $d$. Under the DOMINO framework, the context encoder inferred two disentangled context vectors: context 0 and context 1.
As shown in Figure \ref{fig: same mass}
and Figure \ref{fig:same damping}, the context 1 is more related to damping. When the confounders are set as the same mass but different damping, the visualization result of context 1 under different settings are separated clearly from each other, while under the same damping but different mass settings, the visualization result of context 1 is much more blurred from each other. Similarly, context 0 is more related to mass. When the confounders are set to the same damping but different mass, the visualization result of context 0 under different settings is separated clearly from each other, while under the same mass but different damping settings, the visualization result of context 0 is less different from each other.
}

\subsection{Visualization of the whole context}
\label{app-vis}
\begin{figure}[htbp]
     \centering
     \begin{subfigure}[b]{0.49\textwidth}
         \centering
         \includegraphics[width=\textwidth]{vis/hopper_vis.pdf}
         \caption{Visualization in Hopper-$m$-$d$}
         \label{fig:y equals x}
     \end{subfigure}
     \hfill
     \begin{subfigure}[b]{0.49\textwidth}
         \centering
         \includegraphics[width=\textwidth]{vis/Cripple_ant_vis.pdf}
         \caption{Visualization in Cripple-Ant-$m$-$d$}
         \label{fig:three sin x}
     \end{subfigure}
     
     \begin{subfigure}[b]{0.49\textwidth}
         \centering
         \includegraphics[width=\textwidth]{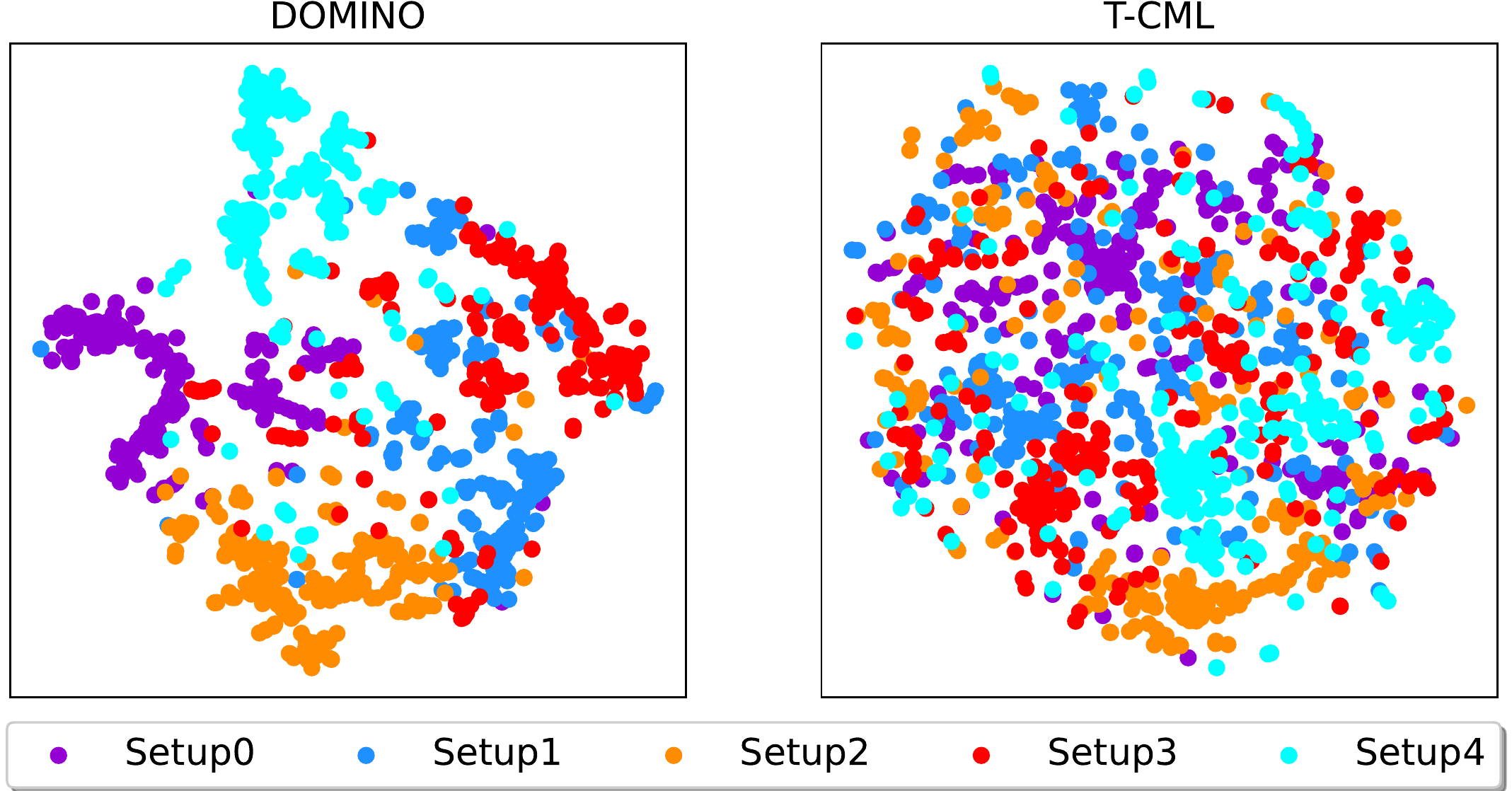}
         \caption{Visualization in Slim-Humanoid-$m$-$d$}
         \label{fig:three sin x}
     \end{subfigure}
     \hfill
     \begin{subfigure}[b]{0.49\textwidth}
         \centering
         \includegraphics[width=\textwidth]{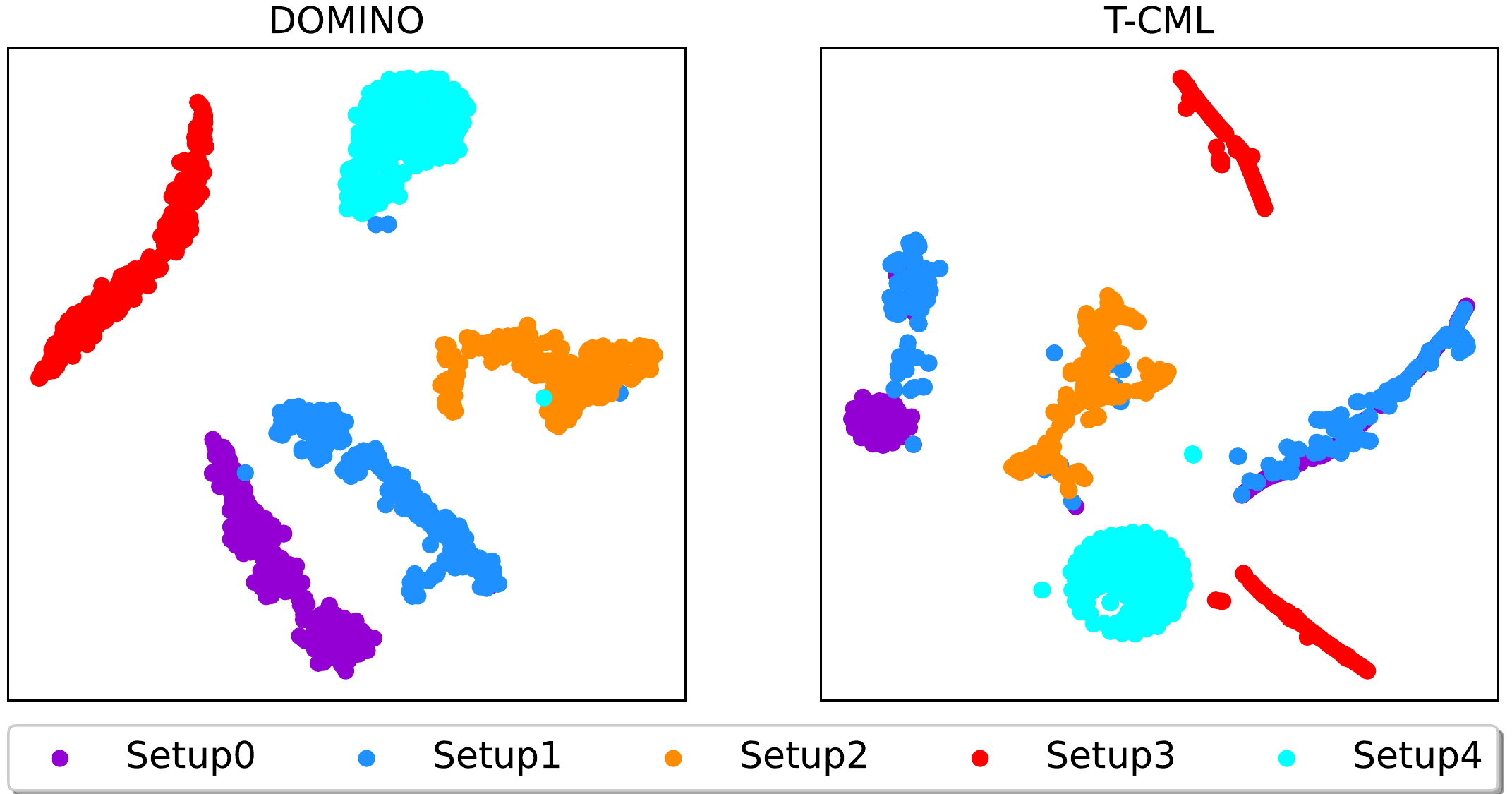}
         \caption{Visualization in Halfcheetah-$m$-$d$}
         \label{fig:three sin x}
     \end{subfigure}
     \hfill
     \begin{subfigure}[b]{0.49\textwidth}
         \centering
         \includegraphics[width=\textwidth]{ 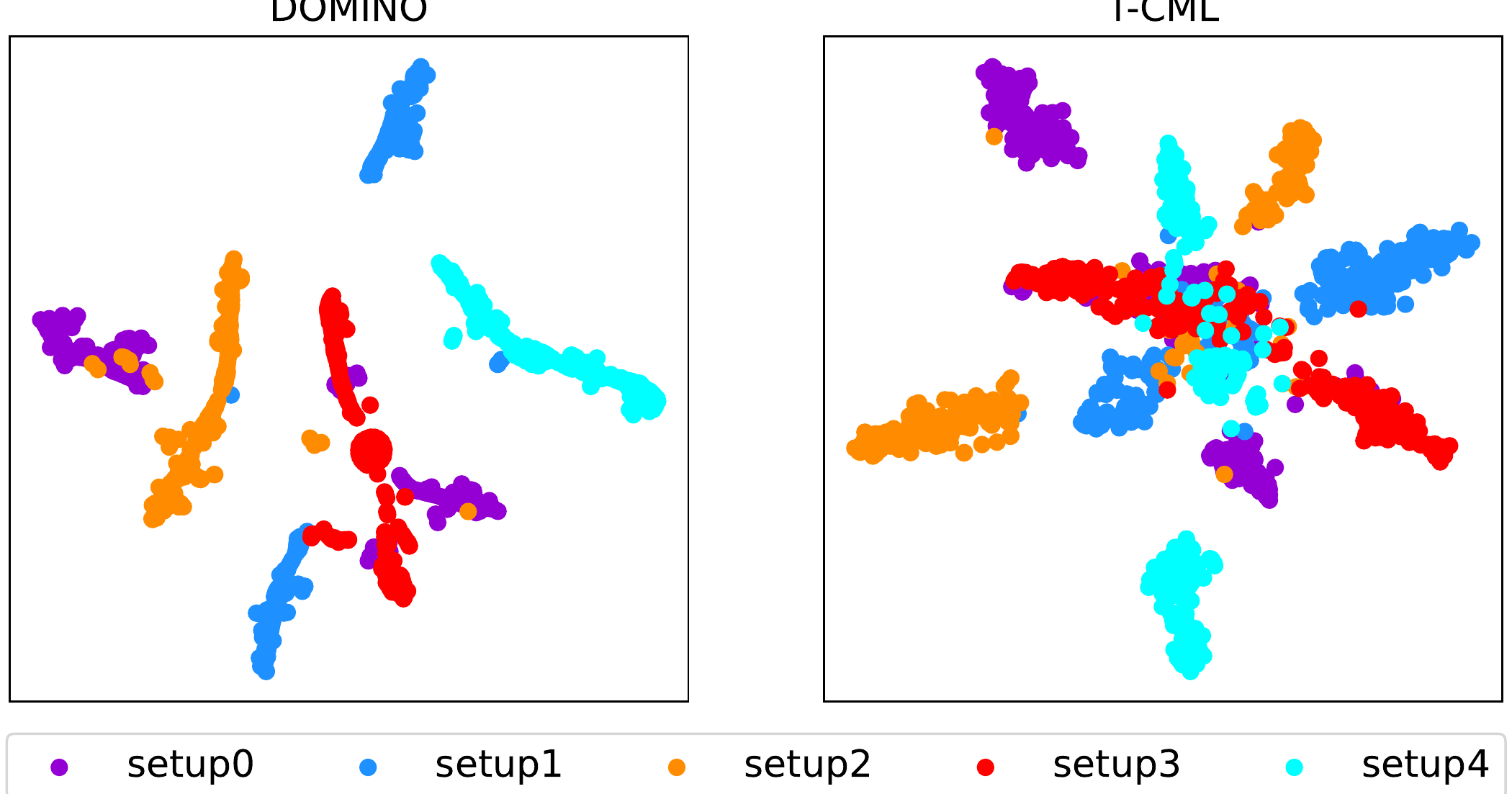}
         \caption{Visualization in Ant-$m$-$d$}
         \label{fig:three sin x}
     \end{subfigure}
     \hfill
     \begin{subfigure}[b]{0.49\textwidth}
         \centering
         \includegraphics[width=\textwidth]{ 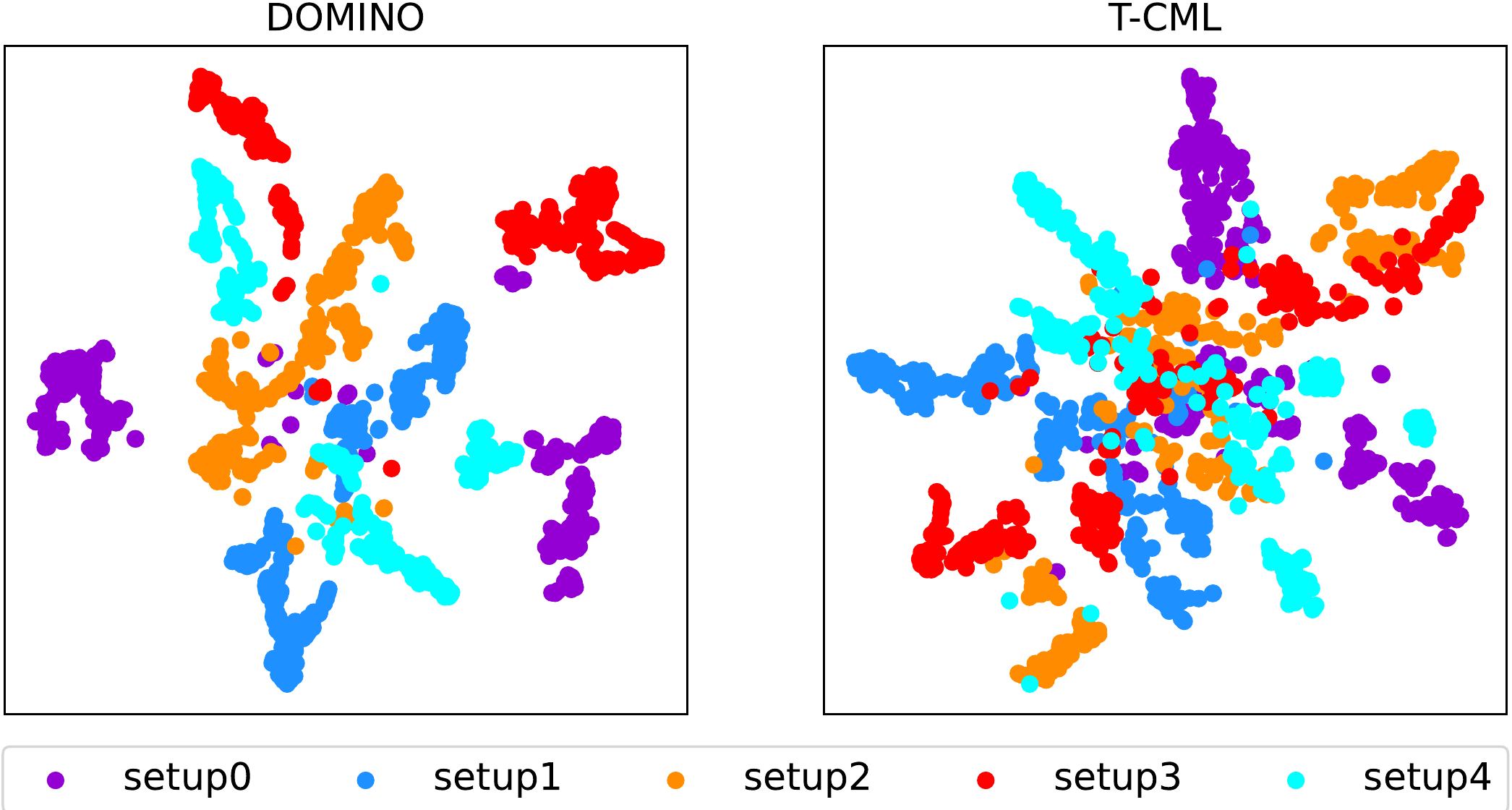}
         \caption{Visualization in Cartpole-$f$-$l$}
         \label{fig:three sin x}
     \end{subfigure}
     \hfill
     \begin{subfigure}[b]{0.49\textwidth}
         \centering
         \includegraphics[width=\textwidth]{ 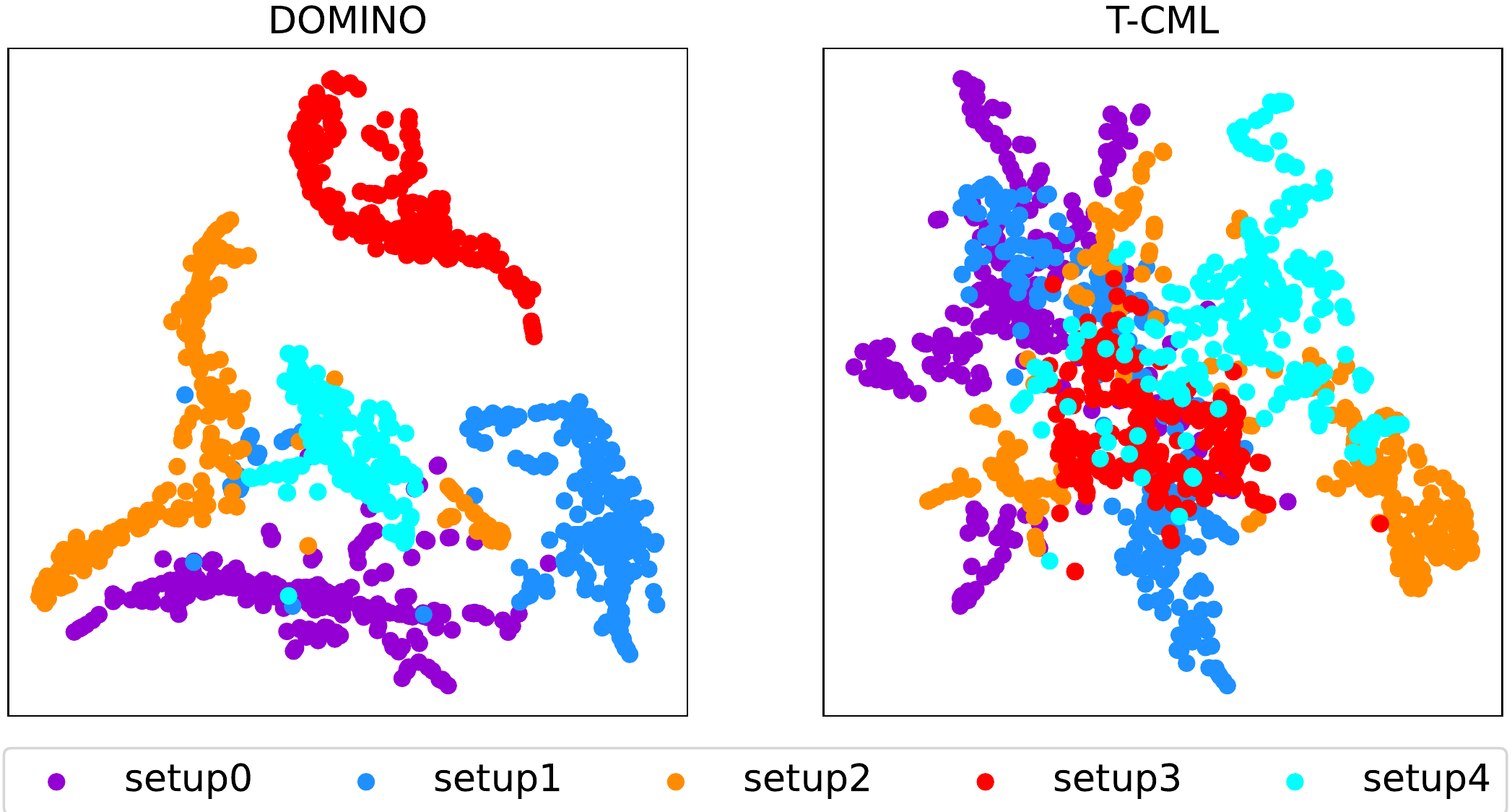}
         \caption{Visualization in Pendulum-$m$-$l$}
         \label{fig:three sin x}
     \end{subfigure}
     \hfill
     \begin{subfigure}[b]{0.49\textwidth}
         \centering
         \includegraphics[width=\textwidth]{ 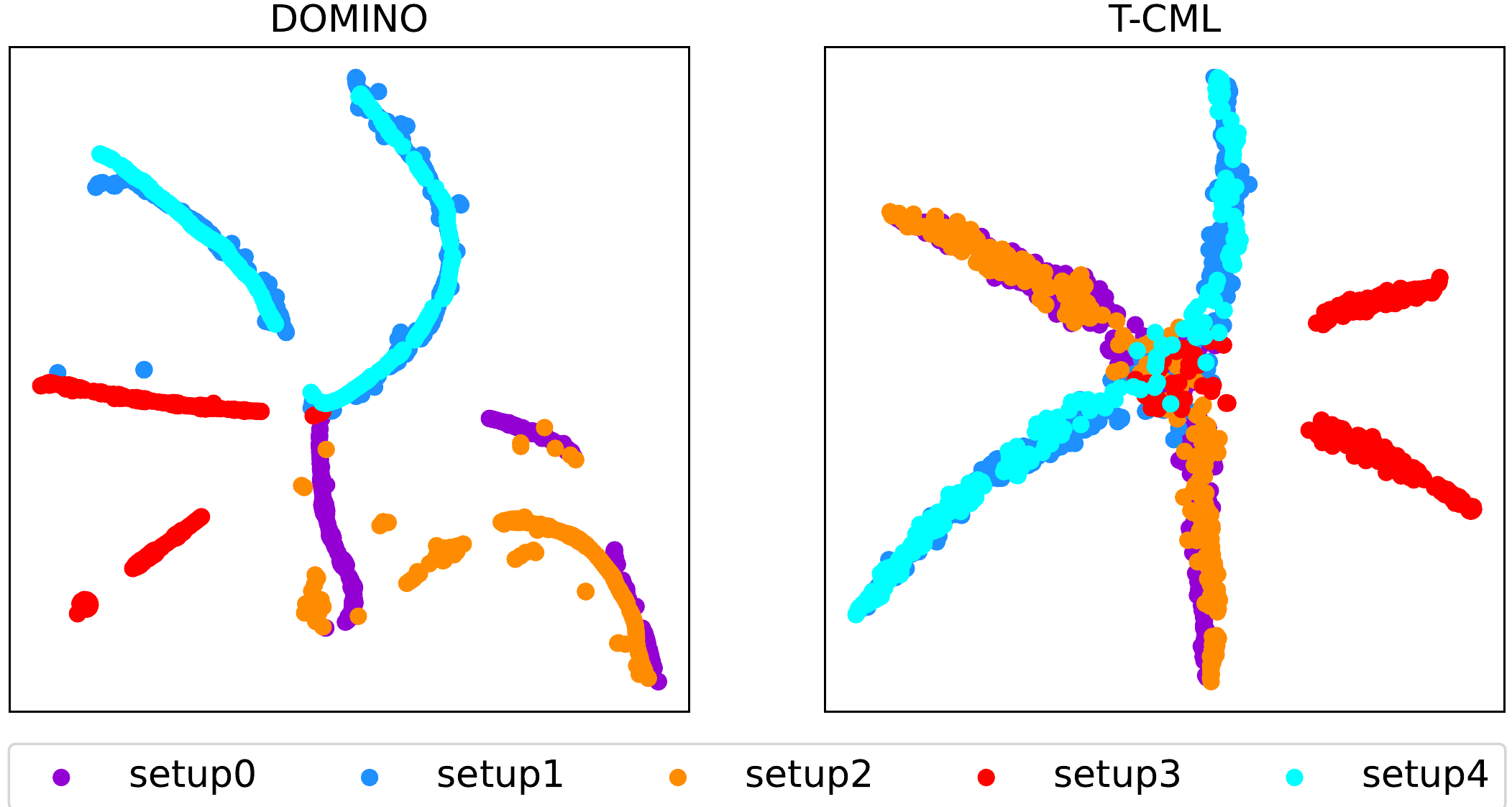}
         \caption{Visualization in Cripple-Halfcheetah-$m$-$d$}
         \label{fig:three sin x}
     \end{subfigure}
     
        \caption{t-SNE \cite{maaten2008visualizing} visualization of context vectors extracted from trajectories collected in various environments. Embedded points from environments with the same confounders have the same color.}
        \label{fig:vis tsne app}
\end{figure}
\begin{figure}[htbp]
     \centering
     \begin{subfigure}[b]{0.49\textwidth}
         \centering
         \includegraphics[width=\textwidth]{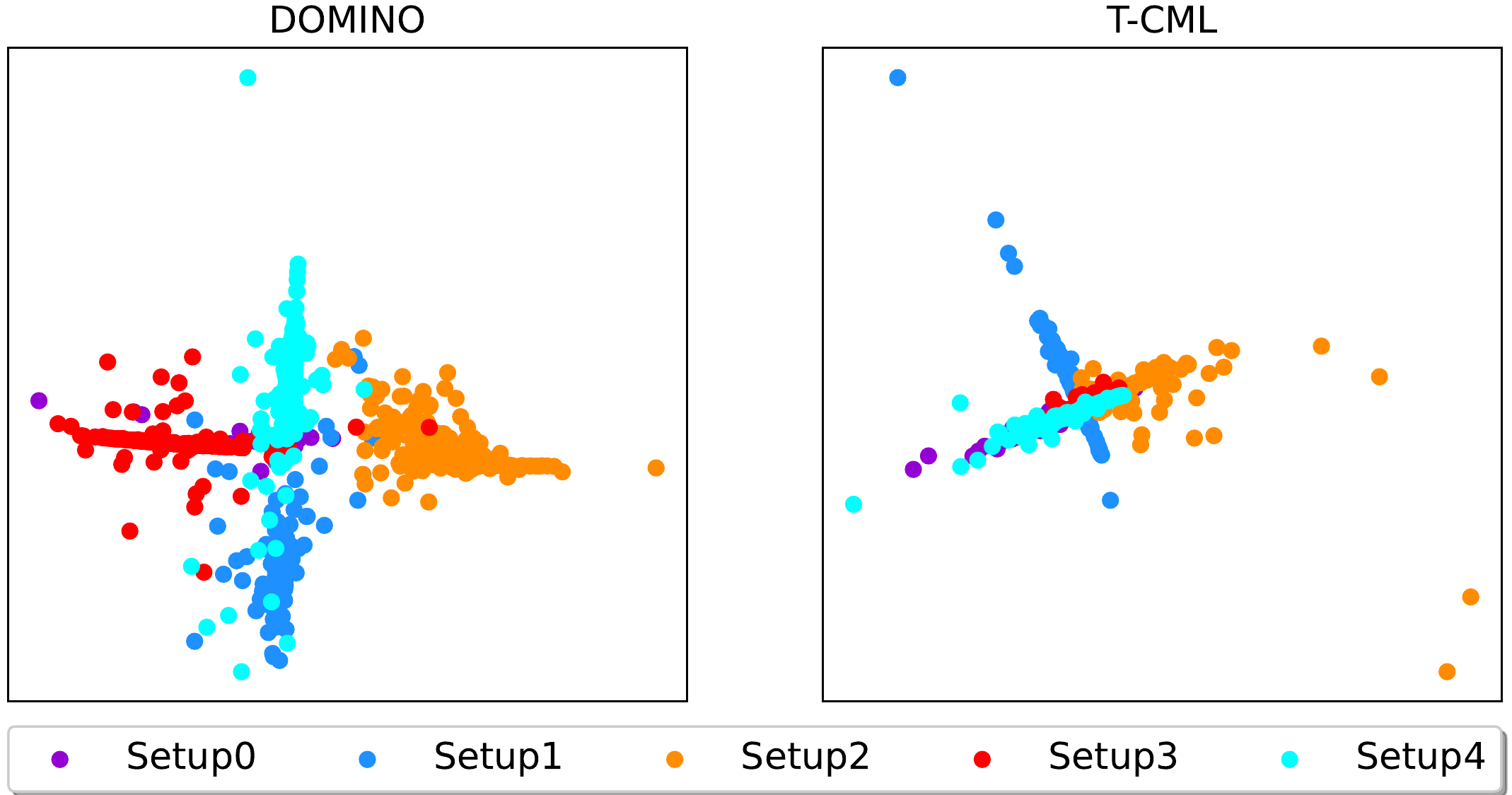}
         \caption{Visualization in Hopper-$m$-$d$}
         \label{fig:y equals x}
     \end{subfigure}
     \hfill
     \begin{subfigure}[b]{0.49\textwidth}
         \centering
         \includegraphics[width=\textwidth]{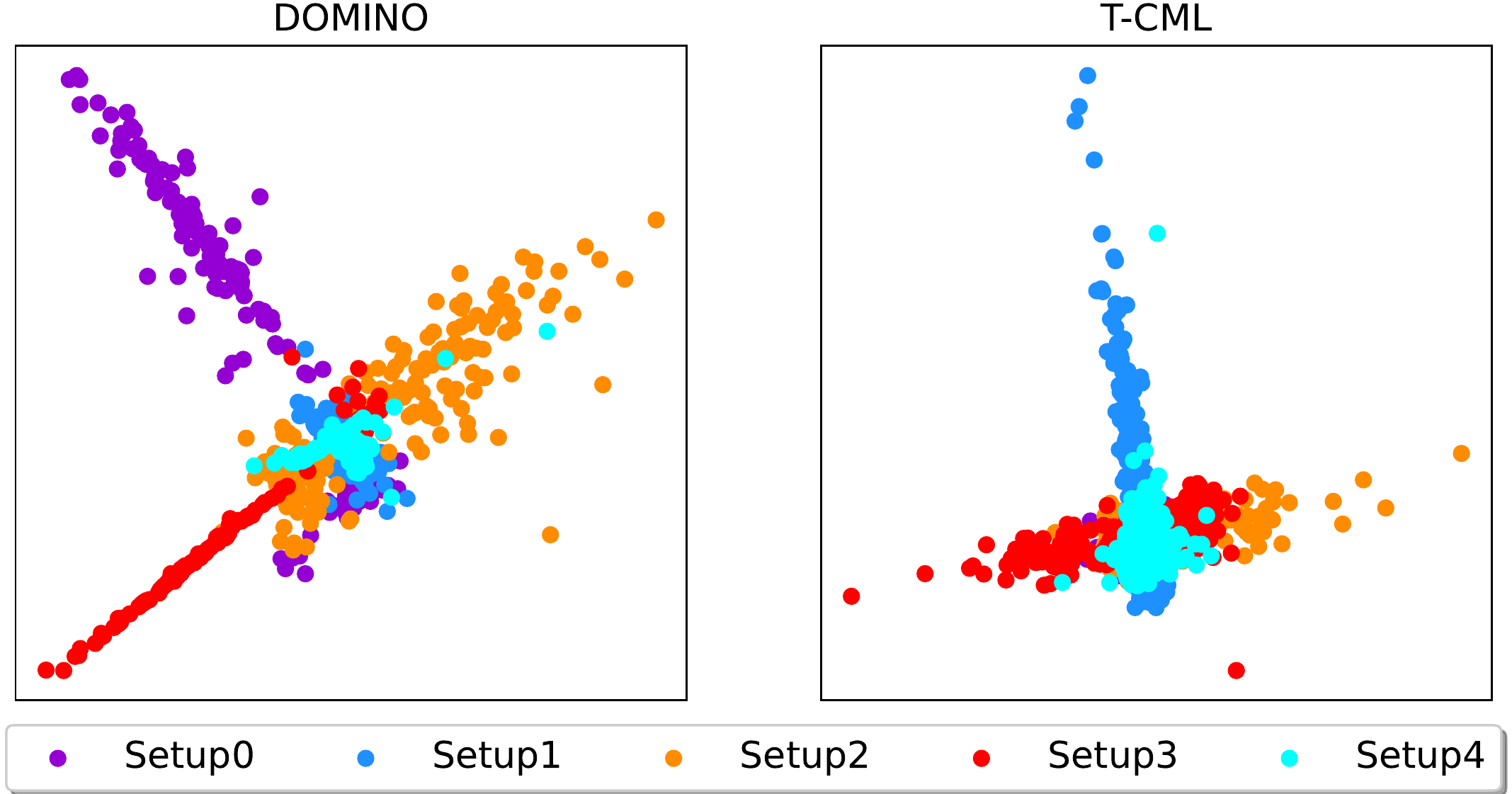}
         \caption{Visualization in Cripple-Ant-$m$-$d$}
         \label{fig:three sin x}
     \end{subfigure}
      \hfill
     \begin{subfigure}[b]{0.49\textwidth}
         \centering
         \includegraphics[width=\textwidth]{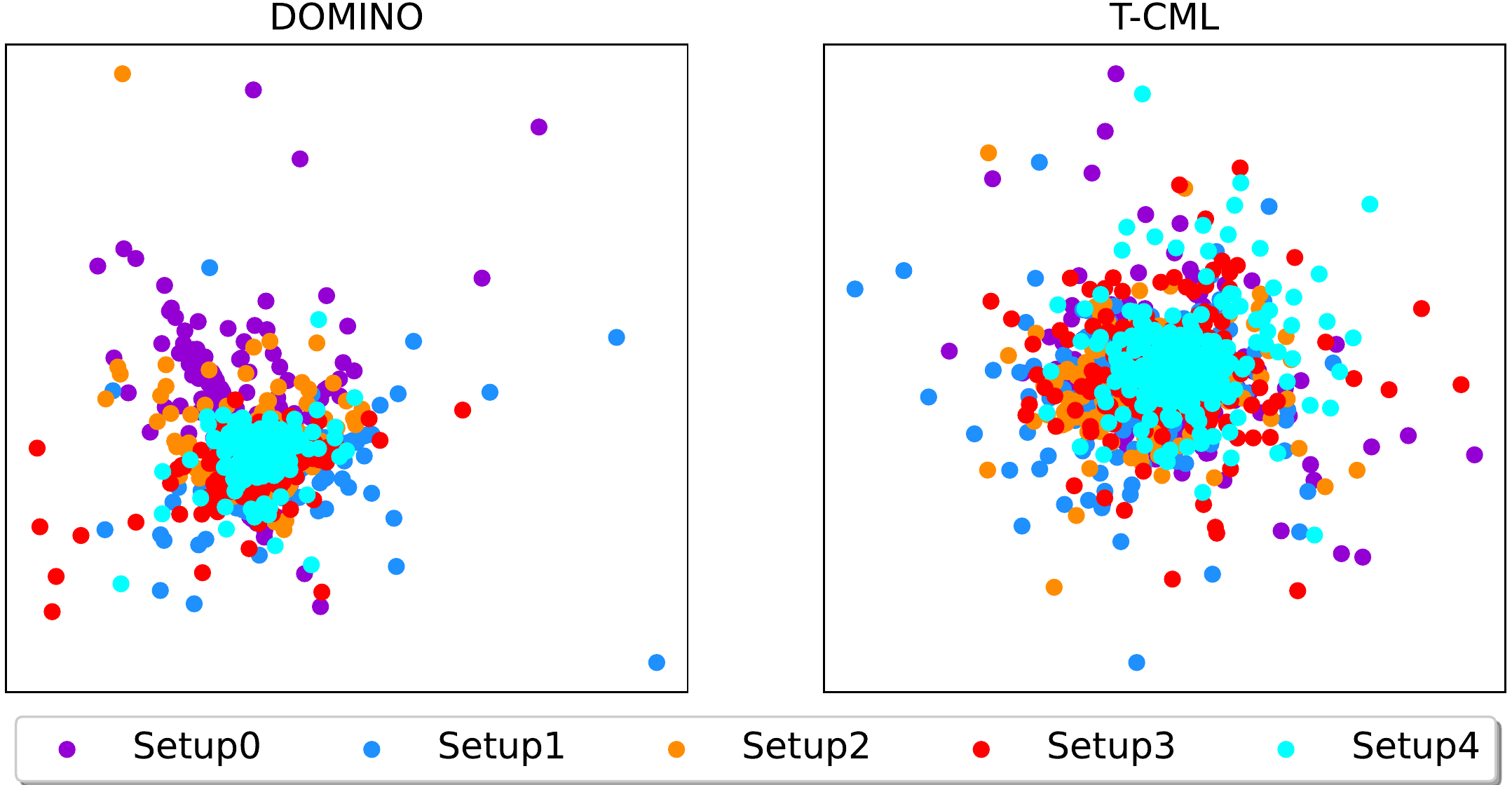}
         \caption{Visualization in Slim-Humanoid-$m$-$d$}
         \label{fig:three sin x}
     \end{subfigure}
      \hfill
     \begin{subfigure}[b]{0.49\textwidth}
         \centering
         \includegraphics[width=\textwidth]{ 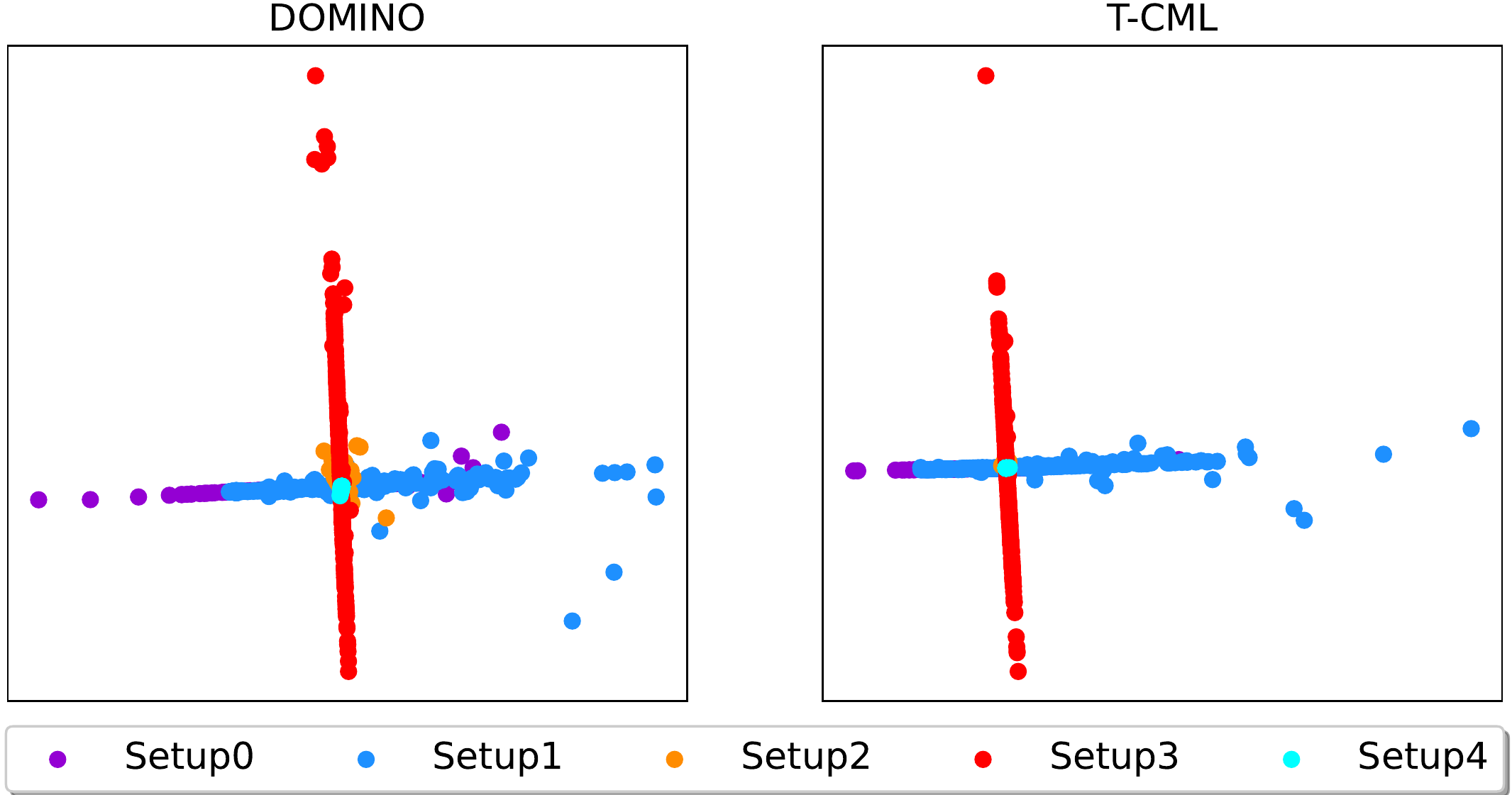}
         \caption{Visualization in Halfcheetah-$m$-$d$}
         \label{fig:three sin x}
    
     \end{subfigure}
     \hfill
     \begin{subfigure}[b]{0.49\textwidth}
         \centering
         \includegraphics[width=\textwidth]{ 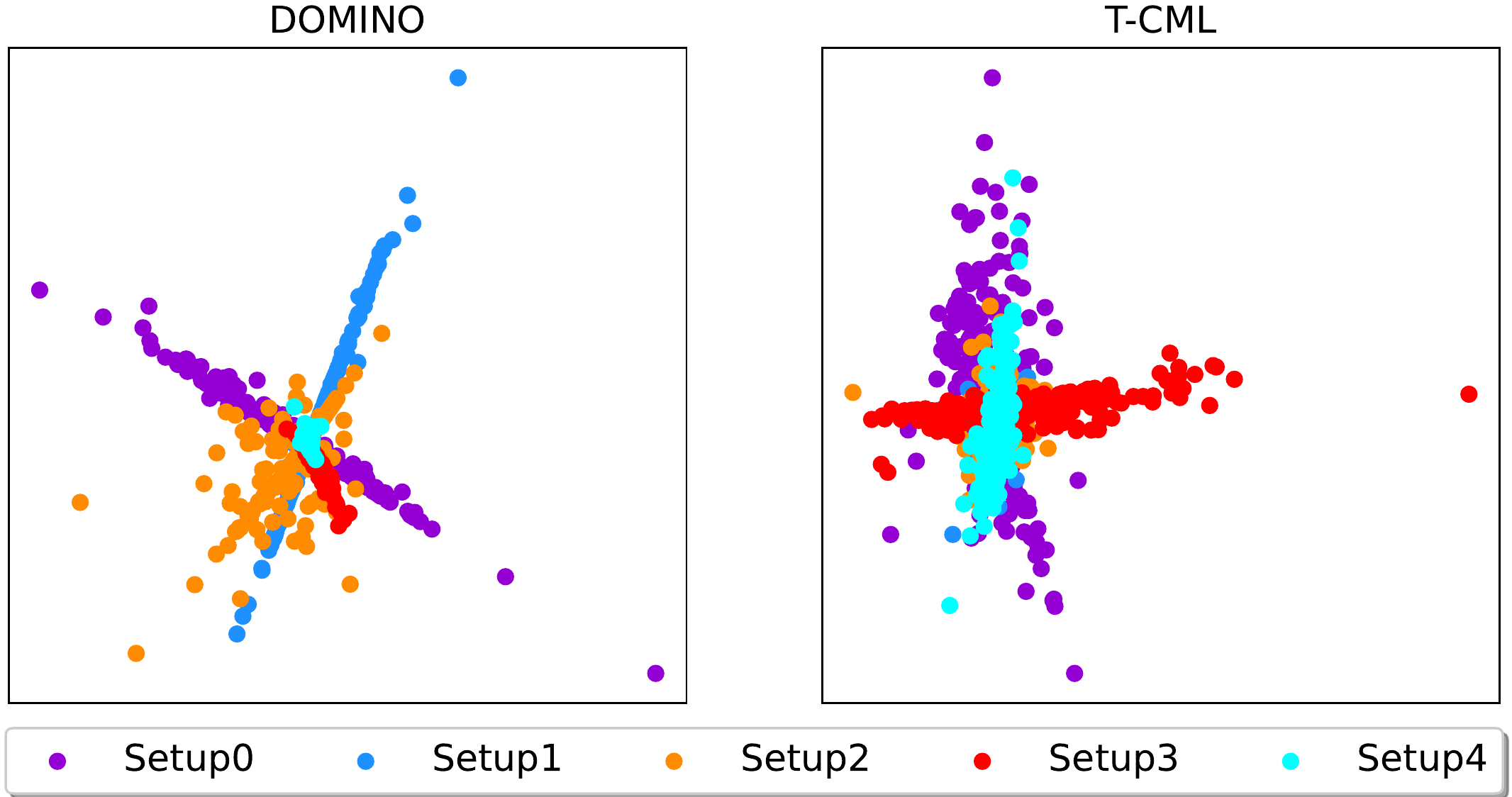}
         \caption{Visualization in Ant-$m$-$d$}
         \label{fig:three sin x}
     \end{subfigure}
     \hfill
     \begin{subfigure}[b]{0.49\textwidth}
         \centering
         \includegraphics[width=\textwidth]{ 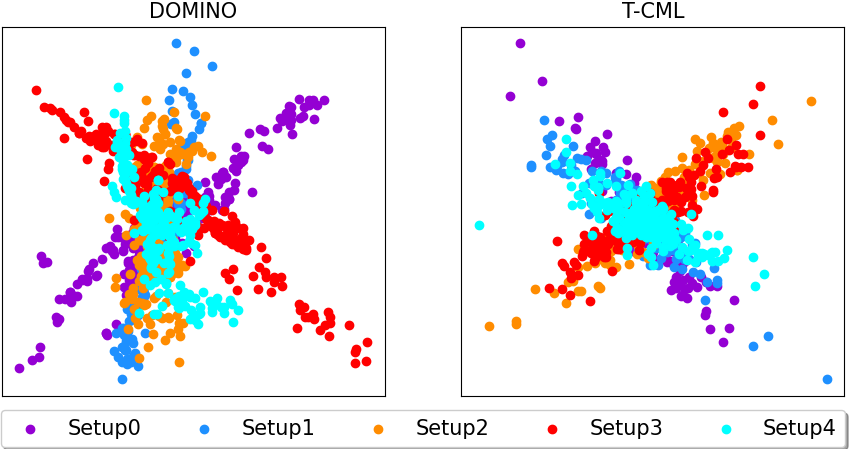}
         \caption{Visualization in Cartpole-$f$-$l$}
         \label{fig:three sin x}
     \end{subfigure}
     \hfill
     \begin{subfigure}[b]{0.49\textwidth}
         \centering
         \includegraphics[width=\textwidth]{ 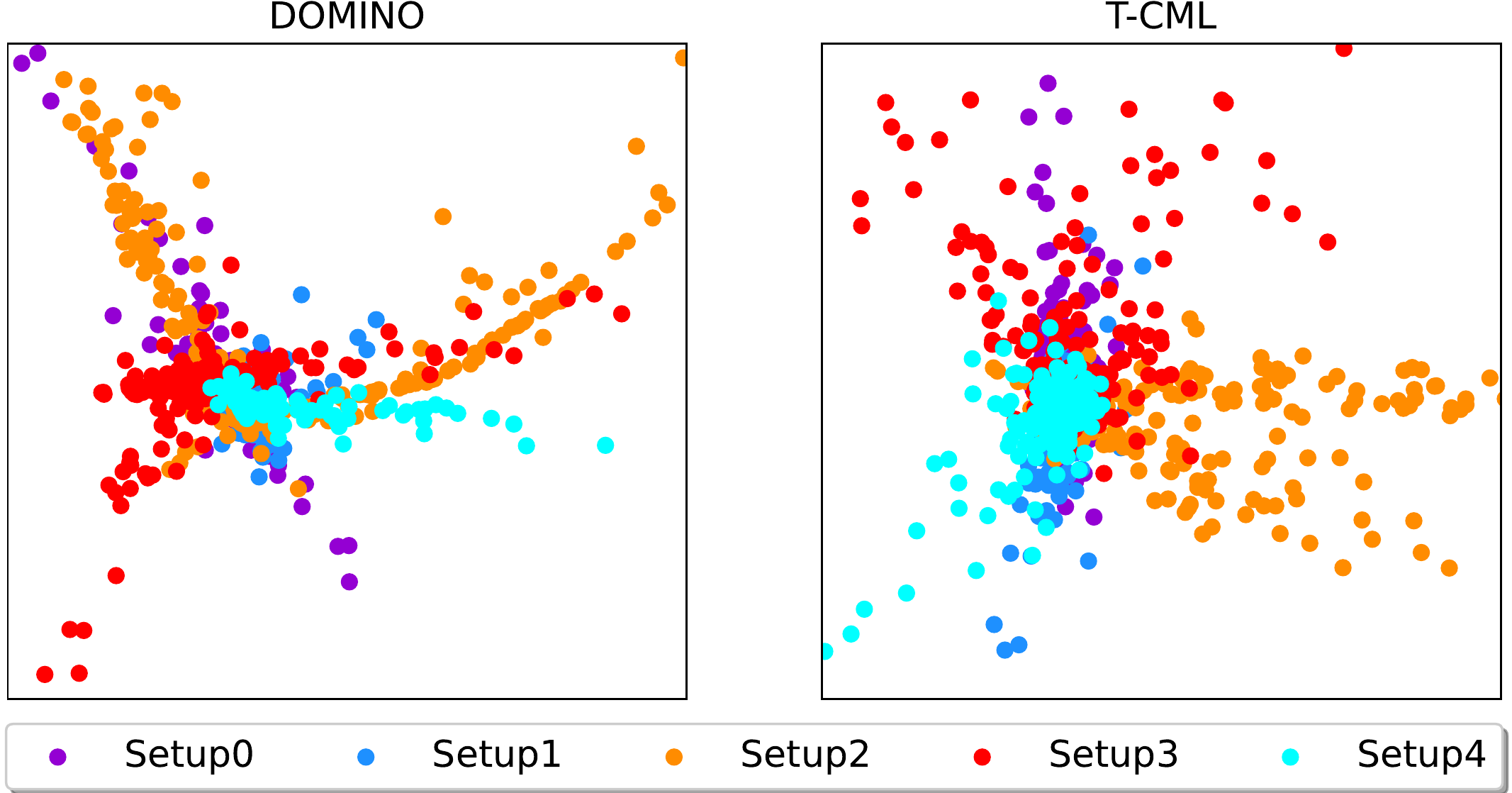}
         \caption{Visualization in Pendulum-$m$-$l$}
         \label{fig:three sin x}
     \end{subfigure}
     \hfill
     \begin{subfigure}[b]{0.49\textwidth}
         \centering
         \includegraphics[width=\textwidth]{ 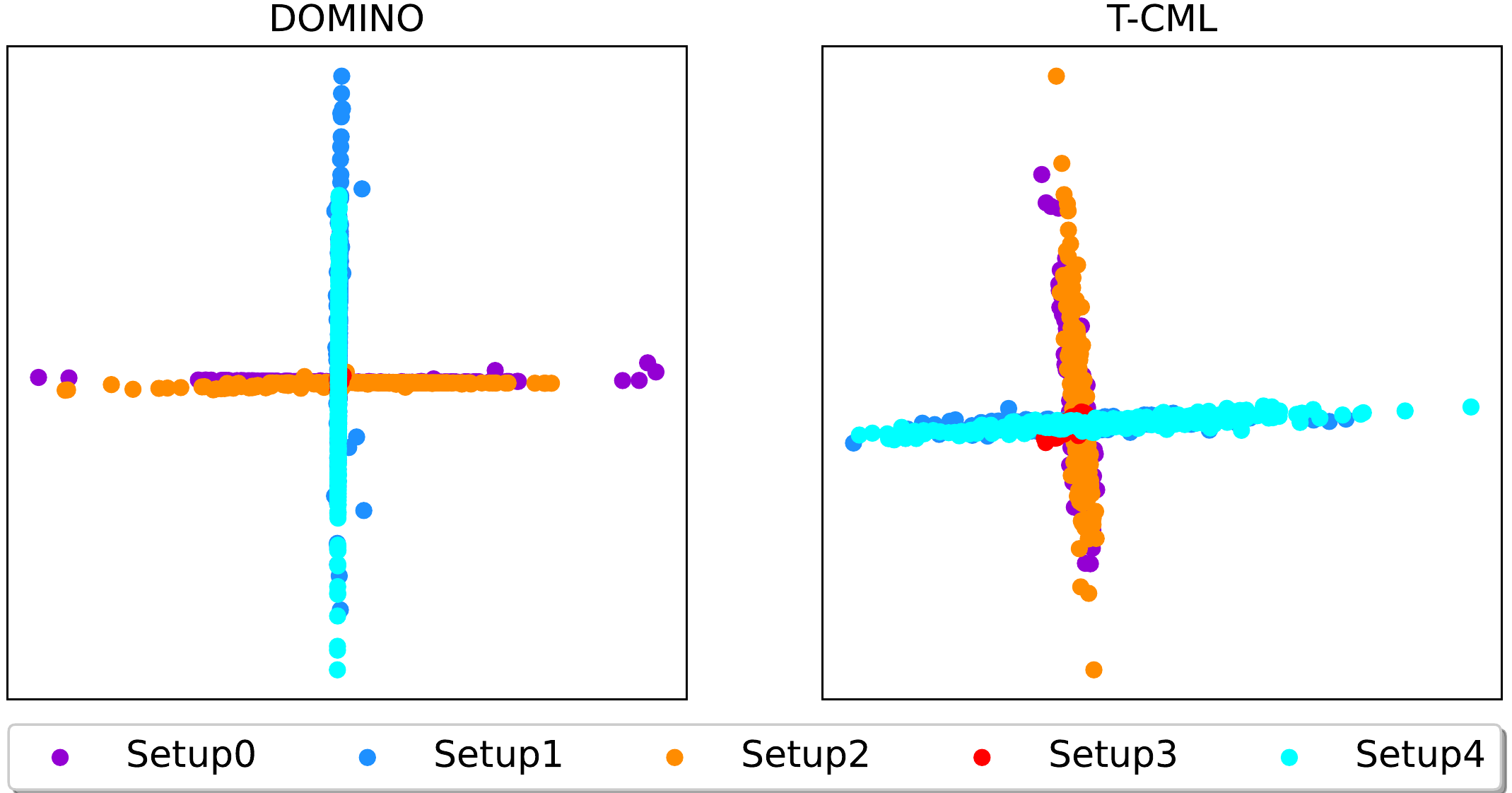}
         \caption{Visualization in Cripple-Halfcheetah-$m$-$d$}
         \label{fig:three sin x}
     \end{subfigure}
        \caption{PCA \cite{maaten2008visualizing} visualization of context vectors extracted from trajectories collected in various environments. Embedded points from environments with the same confounders have the same color.}
        \label{fig:vis pca app}
\end{figure}

\textbf{Visualization}.   { We visualize the whole context which is a  the concatenation of the disentangled contexts learned by \modelname via t-SNE \cite{maaten2008visualizing} and compare it with the entangled context learned by T-MCL.} We run the learned policies under 5 randomly sampled setups of multiple confounders and collect 200  trajectories for each setting. Further, we encode the collected trajectories into context in embedding space and visualize via t-SNE \cite{maaten2008visualizing}  and PCA \cite{jolliffe2002principal}. As shown in Figure \ref{fig:vis tsne app} and Figure \ref{fig:vis pca app}, we find that the disentangled context vectors encoded from trajectories collected under different confounder settings 
could be more clearly distinguished in the embedding space than the entangled context learned by T-MCL.
This indicates that DOMINO extracts high-quality task-specific information from the environment compared with T-MCL. 
Accordingly, the policy conditioned on the disentangled context is more likely to get a higher expected return on dynamics generalization tasks, which is consistent with our prior empirical findings.

\newpage

\section{Further discussion about the future works}

\subsection{Expand DOMINO into reward generalization}
  {
The reward generalization can be categorized as a kind of task generalization.
The parameter of the reward function, for example, the target speed of the robot, can also be considered as a confounder that influences the reward transition.
To address this problem under the DOMINO framework,  we provide the following solution. 
The context encoder maps the current sequence of state-action-reward pairs $\{s_{\tau},a_{\tau},r_{\tau}\}^{t}_{t-H}$ into disentangled contexts, which contains the information of the physical confounders like mass and damping and the reward confounder. The historical trajectory also should consider the reward part, i.e., $s_{t},a_{t},r_{t},s_{t+1}$. Then the proposed decomposed mutual information optimization method can also be used in this situation to extract effective context. Moreover, the prediction loss should also add the reward prediction term. Thus, with the above design, DOMINO can address the reward generalization and dynamics generalization simultaneously.}

\subsection{Expand DOMINO to support related confounders}
  {
To further support the complex environment with confounders related to each other, we can explore how to extract the information that is most useful for state transfer from each of the confounders separately when they do have some correlation with each other. One possible option is to adjust the penalty factor for mutual information between the context vectors in DOMINO, which can be set to be dynamically adjustable.}

\subsection{Combined with VariBad and RIA}
   {VariBad \cite{zintgraf2019varibad} introduces the VAE method and recurrent network to learn the context, which optimizes the context learning from different perspectives from DOMINO and TMCL methods. We believe the effective combination of DOMINO and Varibad will become a more powerful baseline for meta-RL.}
  {
RIA\cite{guo2021relational} doesn't need to record if the two trajectories are collected in the same episode, since the relational intervention approach could optimize the mutual information without environment labels and even without the environment ID, which provides a promising direction of unsupervised dynamics generalization.
We believe that DOMINO and RIA are not in competition, on the contrary, their effective combination will become a stronger baseline, for example, the decomposed MI optimization can be expanded into the relational intervention approach proposed in RIA.}

\end{document}